\documentclass[10pt,article]{IEEEtran}

\usepackage[nocompress,noadjust]{cite}

\usepackage{graphicx}
\usepackage{epsfig}
\usepackage{epstopdf}
\usepackage{multirow}
\usepackage{flushend}
\usepackage{array}
\usepackage{type1cm}
\usepackage{booktabs}
\usepackage{array}
\newcolumntype{P}[1]{>{\centering\arraybackslash}p{#1}}
\newcolumntype{M}[1]{>{\centering\arraybackslash}m{#1}}

\usepackage[justification=centering]{caption}
\usepackage{amsmath}
\usepackage{url}

\usepackage[T1]{fontenc}
\usepackage{microtype}

\usepackage{subfigure,amssymb,amsmath,epsfig,fancyheadings}
\usepackage{epstopdf}
\usepackage{multirow}
\usepackage{booktabs}
\usepackage{comment}

\usepackage{graphicx}
\usepackage{epsfig}
\usepackage{epstopdf}
\usepackage{multirow}
\usepackage{flushend}
\usepackage{array}
\usepackage{type1cm}
\usepackage{array}
\usepackage{booktabs}
 \usepackage{tabularx}
\newcolumntype{P}[1]{>{\centering\arraybackslash}p{#1}}
\newcolumntype{M}[1]{>{\centering\arraybackslash}m{#1}}
\usepackage[figuresright]{rotating}

\usepackage[T1]{fontenc}
\usepackage{microtype}
\usepackage[colorlinks,linkcolor=blue]{hyperref}

\usepackage{url}
\usepackage{colortbl}
\usepackage{arydshln}
\usepackage{multirow}
\usepackage{multicol}
\usepackage{algorithm}
\usepackage{algorithmic}

\usepackage{times}  
\usepackage{helvet} 
\usepackage{courier}  
\usepackage{graphicx} 
\usepackage{amssymb}
\usepackage{amsmath,bm}
\usepackage{subfigure}
\usepackage{multirow}
\usepackage{CJK}
\usepackage[marginal]{footmisc}

\begin{document}
\pagestyle{empty}

\title{Brain-inspired Search Engine Assistant \\ based on Knowledge Graph}


\author{\IEEEauthorblockN{Xuejiao Zhao$^{1,2}$, Huanhuan Chen$^3$, Zhenchang Xing$^4$, Chunyan Miao$^{1,2,*}$\thanks{\noindent* Corresponding author}}\\
\IEEEauthorblockA{$^1$Joint NTU-UBC Research Centre of Excellence in Active Living for the Elderly (LILY), NTU, Singapore\\
$^2$School of Computer Science and Engineering, NTU, Singapore\\
$^3$School of Computer Science and Technology University of Science and Technology of China (USTC)\\
$^4$Research School of Computer Science, Australian National University, Australia\\
Email:\{xjzhao, ascymiao\}@ntu.edu.sg  hchen@ustc.edu.cn  zhenchang.xing@anu.edu.au
}\vspace{-5mm}}
\vspace{-100 mm} 
%

\maketitle
\thispagestyle{empty}
\begin{abstract}
Search engines can quickly response a hyperlink list according to query keywords.
However, when a query is complex, developers need to repeatedly refine the search keywords and open a large number of web pages to find and summarize answers.
Many research works of question and answering~(Q\&A) system attempt to assist search engines by providing simple, accurate and understandable answers. However, without original semantic contexts, these answers lack explainability, making them difficult for users to trust and adopt.
In this paper, a brain-inspired search engine assistant named \emph{DeveloperBot} based on knowledge graph is proposed, which aligns to the cognitive process of human and has the capacity to answer complex queries with explainability.
Specifically, \emph{DeveloperBot} firstly constructs a multi-layer query graph by splitting a complex multi-constraint query into several ordered constraints.
Then it models the constraint reasoning process as subgraph search process inspired by the spreading activation model of cognitive science.
In the end, novel features of the subgraph will be extracted for decision-making.
The corresponding reasoning subgraph and answer confidence will be derived as explanations.
The results of the decision-making demonstrate that \emph{DeveloperBot} can estimate the answers and answer confidences with high accuracy.
We implement a prototype and conduct a user study to evaluate whether and how the direct answers and the explanations provided by \emph{DeveloperBot} can assist developers' information needs.



\end{abstract}

\section{Introduction}
Traditional search engines~(e.g., Google, Bing) provide a list of web sites in which the needed information may be found for developers.
Recently, the community is increasingly recognizing that traditional retrieval models are insufficient to satisfy complex information needs,
and propose to provide additional enhanced services for complex search tasks depending on their characteristics~\cite{sarrafzadeh2014exploring,carlson2003information,singer2013ordinary}.
This is a consensus for major search engine companies (e.g. Google, Bing, Yahoo, etc.): automatic question and answer~(Q\&A) system~(also known as a direct answer search engine, or natural language search engine, etc.) is a more advanced next-generation search engine that returns simple, direct and real-time answers instead of a ranked list of hyperlinks~\cite{Google,Google2,Bing,zhao2020condition,wu2015knowledge}.
However, without the original semantic context, the answers generated by existing Q\&A systems lack explainability that makes the answers difficult for users to trust and adopt~\cite{hong2012multimedia,olvera2013searching,quarteroni2007chatbot,varges2006interactive}.

Knowledge graphs are semantic networks that contain a large number of concepts and relations, which make explainable Q\&A system possible~\cite{tu2000architecture,wu2017knowledge}.
To further provide a human-centred explanation, artificial intelligence~(AI) systems should align with the cognitive model of human and explain within the basic framework of human cognition~\cite{chari2020foundations,yanghua,maheshwari2019learning,murphy2004big,miller2019explanation,high2012era}.
But existing Q\&A systems lack a unified framework of the Q\&A cognitive process based on knowledge graph.
To bridge this gap and equip Q\&A systems with human-like cognitive capabilities, the contents of cognitive science related to Q\&A are explored~\cite{giguere2013limits,matlin1992sensation,owen1997cognitive,collins1975spreading}.

These literature show that the cognitive process of Q\&A is consist of the following five steps including perception, planning, reasoning, response and learning, respectively.
Depending on these inspirations, a brain-inspired search engine assistant called \emph{DeveloperBot} is presented in this paper.
\emph{DeveloperBot} proposes five modules aligned with the human cognitive process of Q\&A.
Its framework can be used as a basis for designing a knowledge graph based Q\&A system that can understand, answer questions and provide a human-centred explanation.

In order to understand the syntax of the technical questions of developers, a rough observation and analysis are made on the closed-ended questions of Stack Overflow.
Close-end questions refer to questions that could be answered with a simple response, e.g., one-word answer.
The results show that these questions are often complex and diverse, and the constraints of a query are often scattered in various grammars, such as attributive clauses, coordinate clause, etc.
Here the constraint consists of three basic elements represented as (category constraint, predicate relation of category constraint and property constraint, property constraint), it describes the information like category, property, etc. of a answer.
However, existing query representation algorithms are insufficient for complex multi-constraint query representation and solving, and the features such as graph topological structure and indirect relations, etc. are not fully utilized in answer reasoning.
To address these issues, two modules of the \emph{DeveloperBot} called \emph{BotPerception} and \emph{BotPlanning} implement a novel query representation algorithm.
Specifically, the \emph{BotPerception} module incorporates the Dependency Parsing into the Tree Parsing~\cite{fellbaum1998semantic} to maximize the completeness of the entities and relations extraction.
Meanwhile, the \emph{BotPlanning} module splits a multi-constraint query into several simple constraints and determines their solving order, then constructs these ordered constraints into a multi-layer query graph for further usage.

Then, a module of the \emph{DeveloperBot} called \emph{BotReasoning} is further proposed to address the complex query solving problem.
This algorithm will use the query graph to search for a candidate subgraph from the knowledge graph by spreading activation algorithm inspired by cognitive science~\cite{collins1975spreading}.
Then, the \emph{BotReasoning} will extract the candidate answers and corresponding features according to direct relation, indirect relation and topological structure of candidate subgraph, etc.
Next, these features will be integrated into the decision making algorithms like DNN~(Deep Neural Network) to determine the final answers.
In the end, a reasoning subgraph and an answer confidence will be extracted following the cognitive process as qualitative and quantitative explanations to explain ``why'', ``how'' and ``how confident'' an answer is being presented.

The experimental results show that, with the novel features of the subgraph, the DNN can extract answers with higher accuracy and estimate answer confidences with lower mean square error~(MSE).
We also implement a prototype of \emph{DeveloperBot} and customize it by loading a knowledge graph of the software engineering domain into its knowledge base.
The results of the user study involving 24 participants show that compared with just using Google, with the assist of \emph{DeveloperBot}, users can find answers faster and with more accuracy.
In addition, using the reasoning subgraph and answer confidence as the explanations of the direct answers can significantly improve the developers' trust and adoption to the answers.
These explanations also assist the developers to understand the answers more deeply, improve the answer accuracy and form better search keywords.
Furthermore, for relatively complex queries, with the assistance of \emph{DeveloperBot}, the search performance improvement of the developers is more significant.


This paper makes the following four major contributions:
\begin{itemize}
  \item [1)] A novel brain-inspired search engine assistant named \emph{DeveloperBot} is proposed, which aligns with the cognitive process of human and have the capacity to answer complex queries with human-centred explainability based on knowledge graph.
  \item [2)] A query representation algorithm implemented by the \emph{BotPerception} and \emph{BotPlanning} modules is proposed, which incorporates the advantages of both Dependency Parsing and Tree Parsing to maximize the completeness of the constraint extraction and determine their solving order, which enhances the representation capacity of the complex multi-constraint query of Q\&A system.
  \item [3)] An algorithm named \emph{BotReasoning} is further proposed for answer reasoning and explanations generation.
      It inspires from spreading activation model and models the constraint reasoning process as candidate subgraph search and decision-making process. Based on the novel features, \emph{BotReasoning} can extract answers with higher accuracy and estimate answer confidences with lower MSE.
  \item [4)] The prototype of the \emph{DeveloperBot} system is implemented and customized by a knowledge graph of software engineering domain as a proof-of-concept.
      A user study is also conducted to evaluate its practical values.
\end{itemize}

\section{Related Works}
Knowledge graph~(KG) has been widely used in Q\&A system, recommendation system and search engines in recent years due to its excellent knowledge representation capacity.
Many research works on recommendation systems have taken advantage of the explainability of knowledge graph to improve the users' trust and adoption prominently~\cite{yao2003web,zhang2018explainable,xu2014two,ma2019jointly}.
Zhang \& Chen summarize explainable recommendation applications in different domains~\cite{zhang2018explainable}, i.e., e-commerce, point of interest, social and multimedia.
Ai et al. present an Indri system that is an explainable recommendation Q\&A system to support modern language technologies~\cite{ai2018learning}.
Catherine, Rose, et al.~\cite{catherine2017explainable} and Yu \& Ren et al.~\cite{yu2014personalized} propose to use knowledge graph entities and meta-path as explanation of recommendations.
There are relatively few research works of Q\&A system explored to improve user trust and adoption by using explainability of knowledge graph.
The content of the explanations of this paper is partially inspired by existing research works of recommendation systems.

The recent research works of Wang \& Zou et al.~\cite{wang2019searching} and Zhu \& Ren et al.~\cite{zhu2015graph} are the two closest works to ours.
Both their works and ours propose to assist the information needs of users expressed in natural language by knowledge graph based Q\&A system.
However, their works extract the entities of a query by simple token matching~\cite{chen2016techland} or entity linking~\cite{wang2019searching,zhu2015graph}.
In contrast, our model analyzes the expression pattern of natural language deeply and construct a query into a multi-layer query graph for further solving, which enhances the representation capacity of the complex multi-constraint query.
Next, although all of these works extract an inference subgraph for further reasoning, this paper extracts the subgraph by spread activation model inspired from cognitive science~\cite{collins1975spreading}, which can extract the candidate answers and corresponding subgraph more comprehensively.
This paper also considers more comprehensive features of the subgraph, such as indirect relation, topological structure, predicate similarity, etc., some of the features are inspired from previous works.
Furthermore, this paper integrates all the features into a decision-making algorithm to extract correct answers with high accuracy.
This also allows our work to quantify the confidence of correct answers.
In addition, our work presents the reasoning subgraph and the confidence as explanations of correct answers, according to the results of the experiments, this significantly improves answer adoption and user confidence.
As far as we know, using Q\&A system with explainability based on knowledge graph as a search engine assistant has not been attempted before.

\section{The Cognitive Framework of \emph{DeveloperBot}}

\begin{figure}[htb]
    \vspace{-0.1cm}
    \centering
    \includegraphics[width=3.5in]{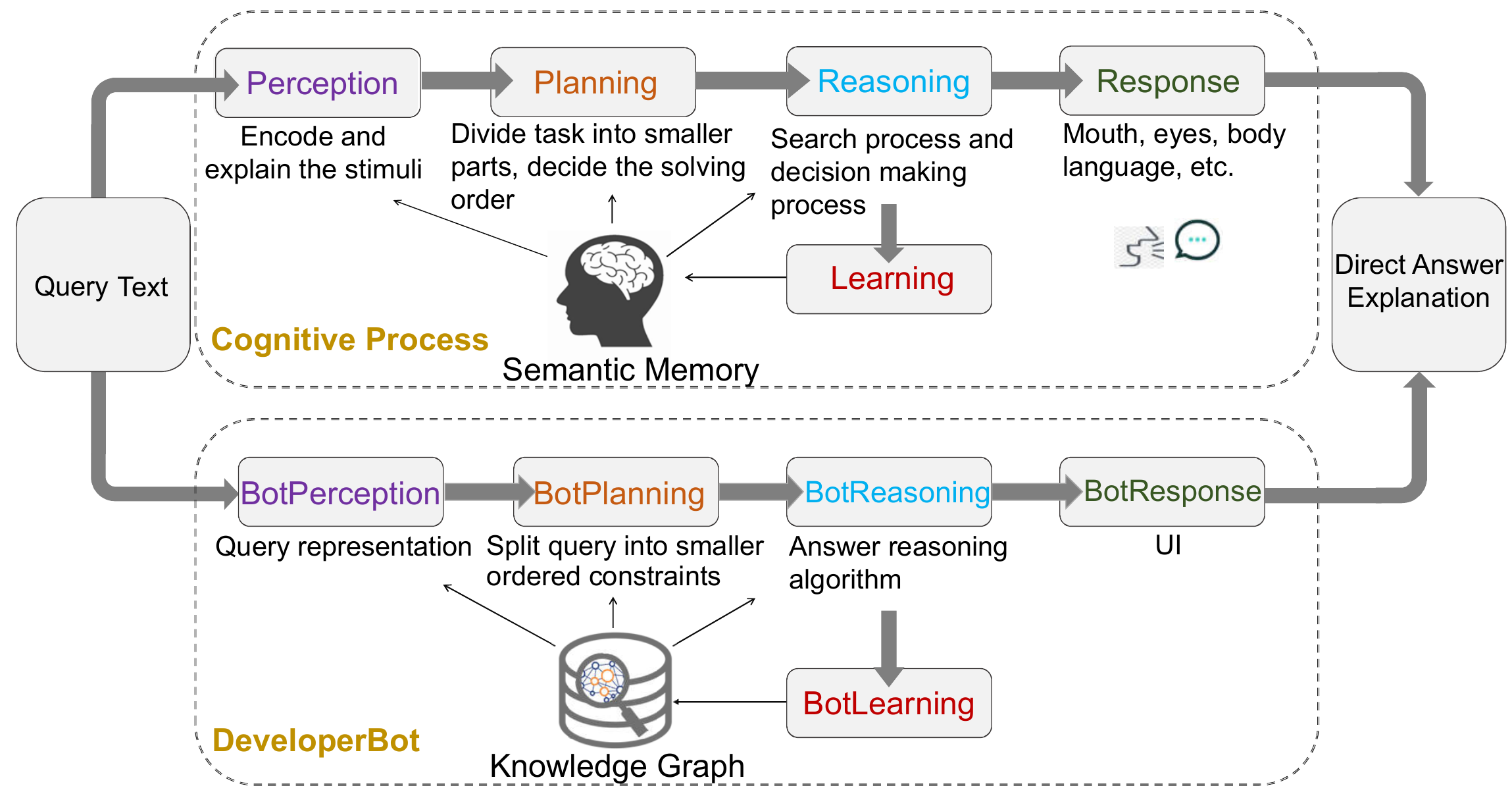}
    \caption {\emph{DeveloperBot} aligned to Cognitive Process}
    \label{cognitiveDeveloperBot}
    \vspace{-0.6cm}
\end{figure}
In the human brain, there are various memory types~(e.g., sensory memory, episodic memory, etc.) that store different contents.
One of them called semantic memory stores the general knowledge of the world, concepts and rules in the form of connected concept nodes.
This is similar to the structure of knowledge graph~\cite{bernecker2011memory,squire1992declarative,quillan1966semantic}.
Question reasoning is an advanced cognitive process of accessing the semantic memory to look for answers according to some premises~\cite{jaya2011knowledge,van2015information,lynch1991memory,giguere2013limits} as shown in the upper half of Fig.~\ref{cognitiveDeveloperBot}.

\begin{figure*}[htb]
    \centering
    \includegraphics[width=7in]{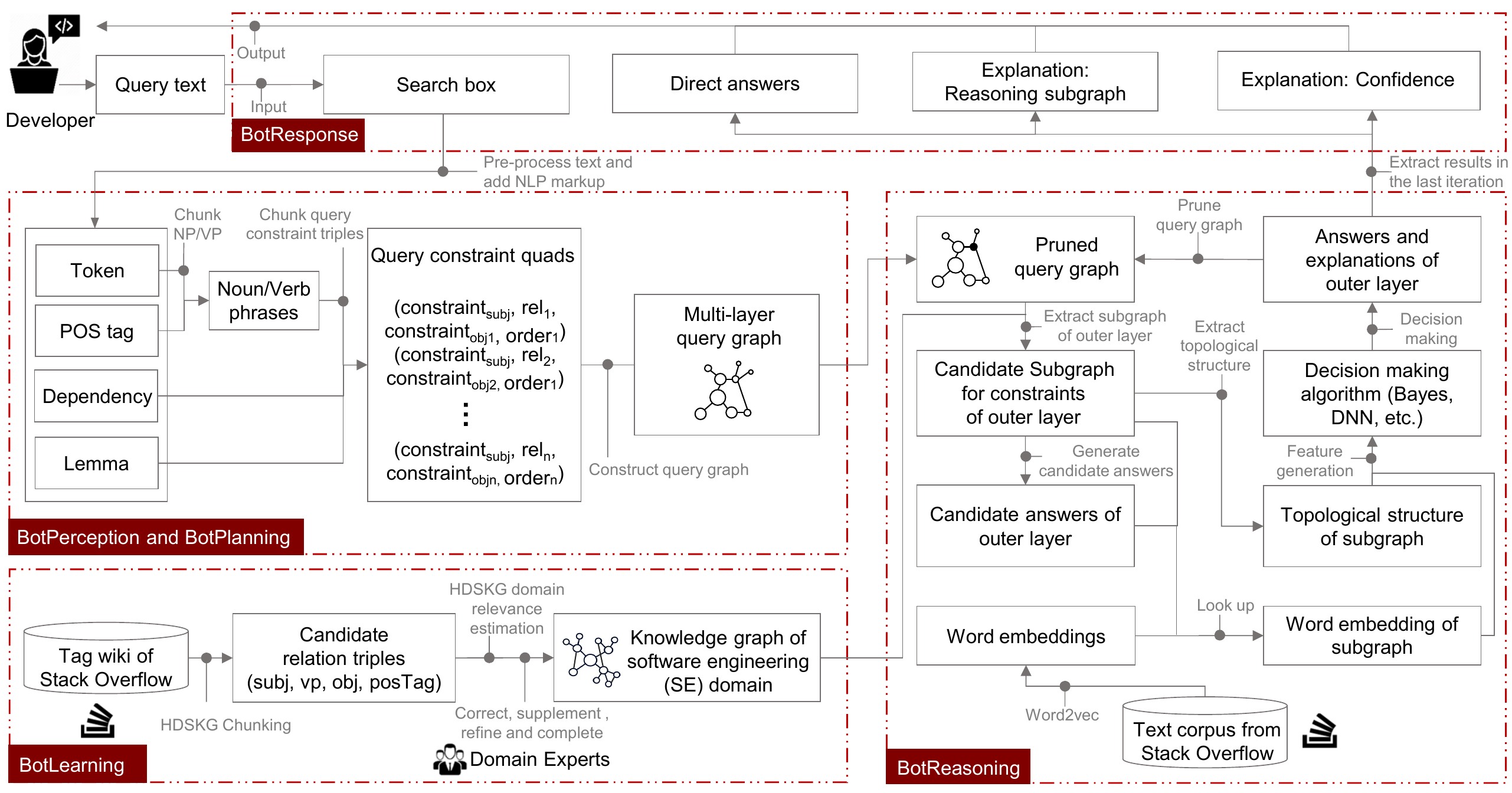}
    \vspace{-0.1cm}
    \caption {An Overview of \emph{DeveloperBot}}
    \label{framework_DeveloperBot}
    \vspace{-0.5cm}
\end{figure*}

Assuming a query as a stimulus from the external environment, we summarize several key steps related to the cognitive process of query reasoning as follows: 1) Perception: to encode and explain external stimuli~(query) as signals that the brain can recognize~\cite{matlin1992sensation}, 2) Planning: to divide a task into smaller, more manageable parts and decide the right executing order~\cite{owen1997cognitive}, 3) Reasoning: to access the semantic memory for the answer~\cite{collins1975spreading}, 4) Response: to output the answers and explanations from the mouth, expression or body language~\cite{olds2019explainable}, 5) Learning: to learn new knowledge by cognitive activities.


The lower half in Fig.~\ref{cognitiveDeveloperBot} is the cognitive framework of \emph{DeveloperBot}.
It shows that how \emph{DeveloperBot} emulates and aligns the cognitive process of the human brain.
The \emph{BotPerception} emulates the perception process for representing the stimuli of the human brain.
The \emph{BotPlanning} algorithm splits a query into smaller constraints and decides their solving order which is similar to the planning of cognitive process.
The \emph{BotReasoning} is answer reasoning based on knowledge graph mimicking the human reasoning~\cite{collins1975spreading}.
The UI~(user interface) called \emph{BotReponse} is similar to the response of the cognitive process.
\emph{BotLearning} is a knowledge graph construction algorithm.
Based on this brain-inspired framework, the human-centred explanations can be generated to show how a query is represented, planed and reasoned.

\section{Approach}

\subsection{An Overview of \emph{DeveloperBot}}
\emph{DeveloperBot} contains four main parts as shown in the overview of Fig.~\ref{framework_DeveloperBot}.
The input is the query text from developers, and there are three outputs:
\begin{itemize}
   \item \emph{Direct answers:} the direct answers derived by our system
   \item \emph{Reasoning subgraph}: a reasoning subgraph to explain the reasons to present each direct answer.
   \item \emph{Confidence}: the confidence of every direct answer.
\end{itemize}
\emph{BotResponse} is an UI that consists of a search box to input query text and an area to visualize direct answers and explanations.
\emph{BotPerception} and \emph{BotPlanning} represent a multi-constraint query as a multi-layer query graph.
\emph{BotReasoning} deduces answers and explanations by the query graph.
\emph{BotLearning} constructs a knowledge graph of software engineering domain offline that combines the knowledge graph from an automatical knowledge graph construction algorithm called \emph{HDSKG} and expertise of the experts in software engineering domain, the detail can be found~\cite{zhao2017hdskg}.

\subsection{\emph{BotPerception} and \emph{BotPlanning}}\label{botPerception}
In this part, we will elaborate on how \emph{BotPerception} and \emph{BotPlanning} combine dependency into constituency-based parse trees to extract, order the constraints in the query and construct them into a multi-layer query graph.

\subsubsection{Annotate Query with Multi-Dimensional NLP Markup}\label{nlpMarkup}
\emph{DeveloperBot} incorporates the tool named \emph{coreNLP} to do the tokenization, POS~(Part of Speech) tagging, dependency parsing and lemmatization to the query~\cite{manning2014stanford,toutanova2000enriching,nivre2016universal,schuster2016enhanced}.
The task of tokenization is to break the sentence into small pieces, called tokens, and drop some characters like punctuation.
POS~(Part of Speech) tagging is the process of labelling words with their grammatical properties, such as NN: Noun, singular or mass, WP: Whpronoun, VBZ: Verb, 3rd person singular present, etc.
Lemmatization can reduce inflectional forms and derivative related forms of a word to a common base form.

\begin{table}[htb]
\vspace{-0.1cm}
\scriptsize
\centering
\caption{Dependencies Description for Sample Query}
\label{Dependency_developerBot}
\begin{tabular}{@{}cccc@{}}
\toprule
Dependency            & Governor                   & Dependent    & \begin{tabular}[c]{@{}c@{}}Semantic relationship between the\\ words depicted by denpendency\end{tabular}                   \\ \midrule
det                   & databases-3                & Which-1      & \begin{tabular}[c]{@{}c@{}}``Which-1'' is determiner of\\ ``databases-3'' \end{tabular}\\ \cdashline{1-4}[0.8pt/2pt]
nsubj                 & support-4                  & databases-3  & \begin{tabular}[c]{@{}c@{}}``databases-3'' is nominal\\ subject of  ``support-3''\end{tabular}                                \\ \cdashline{1-4}[0.8pt/2pt]
dobj                  & support-4                  & Python-5       & \begin{tabular}[c]{@{}c@{}}``Python-5'' is direct objects\\ of ``support-4''\end{tabular}   \\ \cdashline{1-4}[0.8pt/2pt]
nmod:through          & accessed-9                & languages-14 & \begin{tabular}[c]{@{}c@{}}``languages-14'' is nominal\\ modifier of  ``accessed-9''\end{tabular}
\\ \cdashline{1-4}[0.8pt/2pt]
conj:and              & support-4                  & accessed-9  & \begin{tabular}[c]{@{}c@{}}``support-4'' and ``accessed-9'' are\\ connected by coordinating\\ conjunction ``and''\end{tabular} \\ \bottomrule
\end{tabular}
\vspace{-0.2cm}
\end{table}

Dependency is the grammatical structure between the Governor and Dependent.
Tab.~\ref{Dependency_developerBot} presents the detailed description of key dependencies of a sample query~(Unless otherwise specified, the sample query in this paper refers to the query: ``Which graph databases support Python and can be accessed through the RDF query languages that support subgraph extraction?'').
As Tab.~\ref{Dependency_developerBot} shows, ``det'' is the abbreviation of determiner.
There are three interrogative determiners: what, which, whose.
``nsubj'' is a NP and acts as a passive syntactic subject of a clause~\cite{de2008stanford}.
For our sample query, nsubj~(support-4, databases-3) means that ``databases-3'' is the syntactic subject of the verb ``support-4''.
In general, the ``nmod'' indicates some further adjunct relation specified by the case.

\subsubsection{Constituency-based Tree Parsing}\label{TreeParsing}
\begin{table}[htb]
\scriptsize
\vspace{-0.2cm}
\centering
\caption{Parsing Expression of Subconstituent}
\label{Parsing Expression of Subconstituents}
\begin{tabular}{@{}cll@{}}
\toprule
Name & \multicolumn{1}{c}{Parsing Expression}                                                                                                                                                                 &  \\ \midrule
INNP            & \begin{tabular}[c]{@{}l@{}}(IN)+(CD)*(DT)?(CD)*(JJ)*(CD)*(VBD$|$VBG)*(NN.*)*(POS)*(CD)*-\\(VBD$|$VBG)*(NN.*)* (VBD$|$VBG)*(NN.*)*(POS)*(CD)*(NN.*)+\end{tabular}
&  \\ \midrule
WHNP            & \begin{tabular}[c]{@{}l@{}}(WDT$|$WP\$)+(CD)*(DT)?(CD)*(JJ)*(CD)*(VBD)*(NN.*)*(POS)*-\\(CD)*(VBD)*(NN.*)*(VBD)*(NN.*)*(POS)*(CD)*(NN.*)+\end{tabular}
&  \\ \midrule
WHVP            & \begin{tabular}[c]{@{}l@{}}(WP$|$WRB\$)+(MD)*(VB.*)+(JJ)*(RB)*(JJ)*(VB.*)?(DT)?(IN*$|$TO*)+\\ (WP$|$WRB\$)+(MD)*(VB.*)+\end{tabular}
&  \\ \midrule
NP              & \begin{tabular}[c]{@{}l@{}}(CD)*(DT)?(CD)*(JJ)*(CD)*(VBD$|$VBG)*(NN.*)*(POS)*(CD)*-\\(VBD$|$VBG)*(NN.*)*(VBD$|$VBG)*(NN.*)*(POS)*(CD)*(NN.*)+\end{tabular}                                                                                            & \\ \midrule
VP              & \begin{tabular}[c]{@{}l@{}}(MD)*(VB.*)+(CD)*(JJ)*(RB)*(JJ)*(VB.*)?(DT)?(IN*$|$TO*)+\\ 
\end{tabular}
  \\ \bottomrule
\end{tabular}
\vspace{-0.2cm}
\end{table}
In this step, a fulltext parsing technique called ``Tree Parsing'' is adapted to segment a sentence into its subconstituents~\cite{grover2006rule,zhang2002text,loper2002nltk,zhu2015graph}.
As shown in Tab.~\ref{Parsing Expression of Subconstituents}, \emph{DeveloperBot} sets four subconstituents called WHNP, WHVP, NP and VP.
Here NP and VP are abbreviations of the noun phrase and verb phrase, respectively.
Subconstituent WHNP represents a phrase beginning with ``which'', ``what'' or ``whose'' followed by a noun phrase~(e.g. Which relational databases).
If a phrase starts with ``who'', ``when'' or ``what'' followed by a verb phrase, WHVP will be the name of this subconstituent~(e.g. Who developed).

\begin{figure}[htb]
\centering
\includegraphics[width=3.5in]{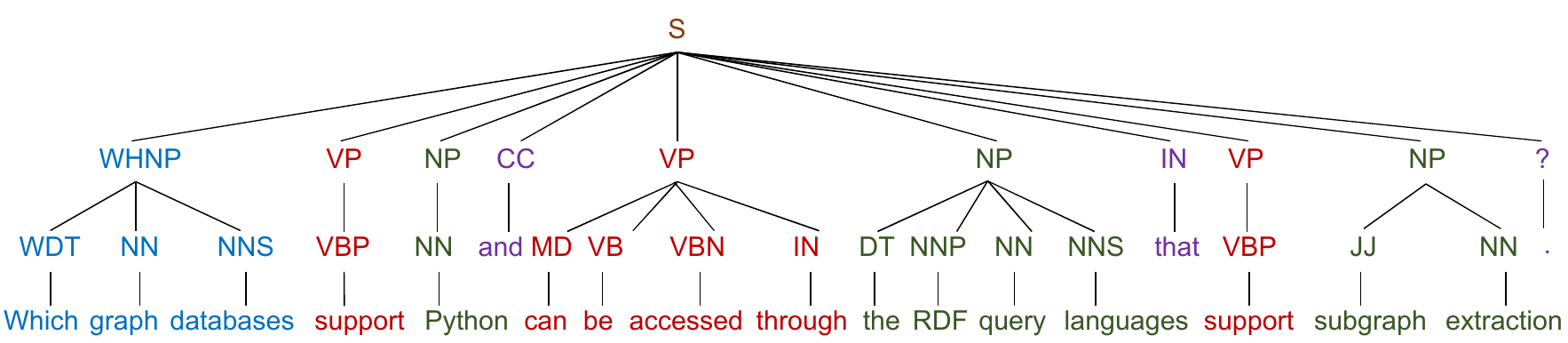}
\vspace{-0.1cm}
\caption{Result of Tree Parsing}
\label{Result of Tree Parsing}
\vspace{-0.6cm}
\end{figure}

The second column of Tab.~\ref{Parsing Expression of Subconstituents} is the parsing expressions of the above four subconstituents.
These parsing expressions are represented by POS tags.
In the parsing expressions, ``?'' stands for whether or not there is such a determinant; ``*'' means zero or more determinant; ``+'' means must have such a determinant; ``-'' means continue to next row.
Using the tree parsing, the tokens of the sentence are segmented to WHNP, WHVP, VP and NP as shown in Fig.~\ref{Result of Tree Parsing}.


\subsubsection{Construct Query Graph with Dependency Parsing}\label{DependencyParsing}
Assume there is a query $Q$, the query graph constructed by \emph{BotPerception} and \emph{BotPlanning} is represented as $QG$.
A $QG$ is composed of the basic units called constraint quads which consists of four basic elements represented as (category constraint, predicate relation of category constraint and property constraint, property constraint, layer number).
The layer number refers to the solving order of this constraint.
A higher layer number means that the constraint is outer layer of $QG$, and the solving sequence is earlier.

Firstly, we use the constituency-based tree parsing to parse $Q$ and get $m$ WHNP, $n$ WHVP, $p$ NP and $q$ VP in $Q$.
Assuming $WHNP_i$, $WHVP_j$, $NP_k$ and $VP_s$ are the $i$th WHNP, $j$th WHVP, $k$th NP and $s$th VP respectively in $Q$:


\begin{equation}\label{vpnp}
\small
\setlength{\abovedisplayskip}{0.05pt}
\setlength{\belowdisplayskip}{2.5pt}
\begin{aligned}
&WHNP_i = (whnp_{i1}, whnp_{i2}, ..., whnp_{if}, ..., whnp_{ix}), \\
&WHVP_j = (whvp_{j1}, whvp_{j2}, ..., whvp_{jw}, ..., whvp_{jy}), \\
&NP_k = (np_{k1}, np_{k2}, ..., np_{kr}, ..., np_{ku}), \\
&VP_s = (vp_{s1}, vp_{s2}, ..., vp_{st}, ..., vp_{sv}),
\end{aligned}
\end{equation}
where $whnp_{if}$, $whvp_{jt}$, $np_{kr}$ and $vp_{st}$ are $f$th, $w$th, $r$th and $t$th word of $WHNP_i$, $WHVP_j$, $NP_k$ and $VP_s$, and the total number of words of them are $x$, $y$, $u$, and $v$, respectively.

Secondly, the dependency parsing technique is used to add dependency markups to all words of $Q$.
After in-depth observation and analysis to the questions on Stack Overflow and our data set, we summarize the query sentence structures commonly used by developers and propose the following 5 query constraints and their layers extraction patterns to construct a query graph.


\textbf{Pattern 1}: If in a query $Q$, $m = 1$ and $n = 0$, and WHNP is detected at the beginning of $Q$.
The NP following the interrogative word called $WHNP_{NP}$ is regarded as a category constraint of direct answers.
Assuming $whnp_{npe}$ is the $e$th word of $WHNP_{NP}$, if the following 4 dependency pairs are detected:
\begin{itemize}
\item $dp[nsubj~(vp_{st}, whnp_{npe}), dobj~(vp_{st}, np_{kr})]$
\item $dp[nsubjpass~(vp_{st}, whnp_{npe}), dobj~(vp_{st}, np_{kr})]$
\item $dp[nsubj~(vp_{st}, whnp_{npe}), nmod~(vp_{st}, np_{kr})]$
\item $dp[nsubjpass~(vp_{st}, whnp_{npe}), nmod~(vp_{st}, np_{kr})]$
\end{itemize}
where $dp[a, b]$ means that $a$ and $b$ appear simultaneously in the result of dependency parsing of $Q$.
Then a property constraints ($VP_s$, $NP_k$) will be extracted into the $QG$.

For the sample query,
\emph{``Which graph databases''} is a WHNP at the beginning of the query phrase and \emph{``graph databases''} is the $WHNP_{NP}$ as shown in Fig.~\ref{Result of Tree Parsing}.
Therefore, (graph\_databases) will be extracted as a category constraint of direct answers.
Then, two dependency pairs $dp[nsubj~(support-4, databases-3)$ and $dobj (support-4, Python-5)]$ and $dp[nsubjpass~(accessed-9, databases-3)$ and $nmod~(accessed-9, languages-14)]$ are detected.
Here ``support-4'' is $vp_{11}$ which means the first word in the first VP of $Q$.
The ``databases-3'' is $whnp_{np2}$ representing the second word of NP of WHNP and so on.

In \emph{BotPerception}, all property constraints belonging to category constraints of direct answers are classified as the innermost layer of $QG$.
Therefore, from the sample query, \textbf{Pattern 1} can extract two constraint quads: (graph\_databases, support, Python, 1) and (graph\_databases, can\_be\_accessed\_through, RDF\_query\_languages, 1)


\textbf{Pattern 2}: Assuming the ``property constraints'' of a constraint quads is $PC = (pc_1, pc_2, ..., pc_g, ..., pc_d)$,
where $pc_g$ is the $g$th word of $PC$, and there are $d$ words in $PC$ in total.

If a constraint quads has been extracted and its layer number equal to $c$.
And then if the following 4 dependency pair are detected:
\begin{itemize}
\item $dp[nsubj~(vp_{st}, pc_g), dobj~(vp_{st}, np_{kr})]$
\item $dp[nsubjpass~(vp_{st}, pc_g), dobj~(vp_{st}, np_{kr})]$
\item $dp[nsubj~(vp_{st}, pc_g), nmod~(vp_{st}, np_{kr})]$
\item $dp[nsubjpass~(vp_{st}, pc_g), nmod~(vp_{st}, np_{kr})]$
\end{itemize}
A constraint quads ($PC$, $VP_s$, $NP_k$, $c+1$) will be extracted into the $QG$.

For the sample query, a constraint quads (graph\_databases, can\_be\_accessed\_through, RDF\_query\_languages, 1) has been extracted, so $PC$ = ``RDF query languages''.
A dependency pair $dp[nsubjpass~(accessed-12, databases-3), nmod~(accessed-9, languages-14)]$ is detected.
Here ``accessed-9'' is $vp_{23}$, ``databases-3'' is $pc_3$, ``languages-14'' is $np_{34}$ as shown in Fig.~\ref{Result of Tree Parsing}.
So the constraint quads (RDF\_query\_languages, support, subgraph\_extraction, 2) will be extracted into the $QG$.



\textbf{Pattern 3}: In a query $Q$,
if $whvp_{11}$ is ``who'' and $WHVP_{VP}$ is not ``is'', ``PERSON'' will be added to $QG$ as a category constraint of direct answers.
For example, a query ``Who created Python?'' will be constructed as (PERSON, created, Python, 1), where the $whvp_{11}$ equals to ``when'', a category constraint ``DATE'' will be added into $QG$.
If $whvp_{11}$ is ``what'' and $WHVP_{VP}$ is not ``is'', a category constraint ANYENTITY~(a wildcard of any entities) will be put into $QG$.


\textbf{Pattern 4}: If in a query $Q$, WHVP is detected at the beginning of the $Q$.
The VP following the interrogative word is ``is'' and the $whvp_{vp1}$ is not ``when'' like query ``What is Java Servlet?''.
We will regard this question as a definition query and construct the $QG$ as (ANYENTITY, ANYRELATION, Java\_Servlet, 1) and (Java\_Servlet, ANYRELATION, ANYENTITY, 1), where ANYRELATION represents a wildcard of any relation.


\textbf{Pattern 5}: If in a query $Q$, $m = 0, n = 0$, and the first word of $Q$ is ``List'' followed by only an NP.
e.g. ``List graph database''.
We will construct the $QG$ as (graph\_database, ANYRELATION, ANYENTITY, 1).

\subsection{\emph{BotReasoning}}\label{BotReasoning}
As described in the~\ref{botPerception}, a query graph $QG$ is constructed.
In this section, we will elaborate on how \emph{BotReasoning} reasons the answers and generates explanations by $QG$.

\subsubsection{Candidate Subgraph Search}\label{subgraphSearch}
Assuming the current constraint quads for candidate subgraph search is $(c_{subj}, c_{predicate}, c_{obj}, c_{layer})$, and the knowledge graph $KG = [ node_1, node_2, ..., node_i, ..., node_n]$, where $node_i$ is the $i$th node of $KG$, $n$ is the nodes number of $KG$.
The pseudo code of subgraph search is shown in Algorithm.~\ref{subgraphSearchAlgorithm}.


Each node in $KG$ has an initial associated activation value $a_i \in \mathbb{R}$ and $0 < a_i < 1$.
A link $link_{ij}$ connects source $node_i$ to target $node_j$ and the weight of $link_{ij}$ is $w_{ij}$ where $w_{ij} \in \mathbb{R}$ and $0 < w_{ij} < 1$.
The nodes of $KG$ have an active threshold $AT$, where $AT \in \mathbb{R}$ and $0 < AT < 1$.
There is a decay factor $DF$, where $DF \in \mathbb{R}$ and $0 < DF < 1$.

\begin{algorithm}[thb]
\small
\setlength{\abovedisplayskip}{1pt}
\setlength{\belowdisplayskip}{0.01pt}
\caption{Subgraph Search}
\begin{algorithmic}[1]\label{subgraphSearchAlgorithm}
\REQUIRE {$(c_{subj}, c_{predicate}, c_{obj}, c_{layer})$: current constraint quads for subgraph search, $KG$: knowledge graph, $ST$: iteration times\\}
\ENSURE $SG$: the searched subgraph, $CR$: the crossover relation
\STATE Define a set $\mathbf{S}$ to store the nodes that can spread activation\\
\STATE Link $c_{subj}$ and $c_{obj}$ to nodes $node_{subj}$ and $node_{obj}$ in $KG$
\STATE Subroutine $SpdActi(node)$
    \STATE Insert $node$ into $\mathbf{S}$
    \WHILE{($ST > 0$)}
        \IF{($\mathbf{S} \neq \emptyset$)}
            \STATE Insert $\mathbf{S}$ into $SG_{temp}$
            \FOR{(each $node_i \in \mathbf{S}$)}
                \IF{($node = node_{obj}$ and $node_j \in SG_{subj}$)}
                   \STATE Insert $(node_j, link_{ij}, node_i)$ into $CR$
                \ELSE
                    \STATE Spread activation and adjust value to every neighbouring $node_j$ according to Equation.\ref{spread} and Equation.\ref{ajpie}
                \ENDIF
            \ENDFOR
            \STATE Replace all the nodes of $\mathbf{S}$ by activated nodes in this round
        \ENDIF
            \STATE $ST--$
    \ENDWHILE
    \RETURN $SG_{temp}$
\STATE EndSubroutine
\STATE $SG_{subj} \gets SpdActi(node_{subj})$   $SG_{obj} \gets SpdActi(node_{obj})$
\STATE $SG \gets SG_{subj} \cup SG_{obj}$
\RETURN $SG_{subj}$, $SG_{obj}$, $SG$ and $CR$
\end{algorithmic}
\end{algorithm}

In the beginning of the subgraph search, \emph{BotReasoning} links the $c_{subj}$ and $c_{obj}$ to the corresponding $node_{subj}$ and $node_{obj}$ in $KG$.
Therefore, the initial activation value of $node_{subj}$ and $node_{obj}$ will be greater than active threshold $AT$.
These two nodes are activated and will initiate to spread the activation to all the neighbouring nodes parallelly.

Assuming $link_{ij}$ connects the source node $node_i$ to the target node $node_j$.
While the activation spread from $node_i$ to $node_j$, $node_i$ will compute $a_{j\_temp}$ as:
\vspace{-2mm}
\begin{equation}\label{spread}
 \begin{aligned}
  a_{j\_temp} = a_j + (a_i * w_{ij} * DF),
 \end{aligned}
\end{equation}

Then, $a_j$ will adjust its value to $a'_j$ according to $a_{j\_temp}$ as following formula:
\begin{equation}\label{ajpie}
a'_j = \left\{
\begin{aligned}
1 & , & if a_{j\_temp} \geq 1, \\
a_{j\_temp} & , & if AT \leq a_{j\_temp} < 1, \\
a_j & , & if a_{j\_temp} < AT.
\end{aligned}
\right.
\end{equation}


Once $node_j$ is activated, it will initiate the next spreading activation cycle to its neighbouring nodes.
The spreading will be terminated under the following three situations:
\begin{itemize}
  \item The spreading arrives the endmost nodes of the $KG$ or exceeds the preset upper bound $ST$.
  \item All the nodes reach $AT$.
  \item During searching $node_{obj}$, if the neighbouring node $node_j$ of $node_i$ belongs to $SG_{subj}$. The spreading will be terminated and $(node_j, link_{ij}, node_i)$ will be inserted into $CR$, where $CR$ are the crossover relations indicating the relations of $c_{subj}$ and $c_{obj}$.
\end{itemize}

\subsubsection{Decision Making}
\begin{figure}[htb]
    \vspace{-0.5cm}
    \centering
    \includegraphics[width=2.6in]{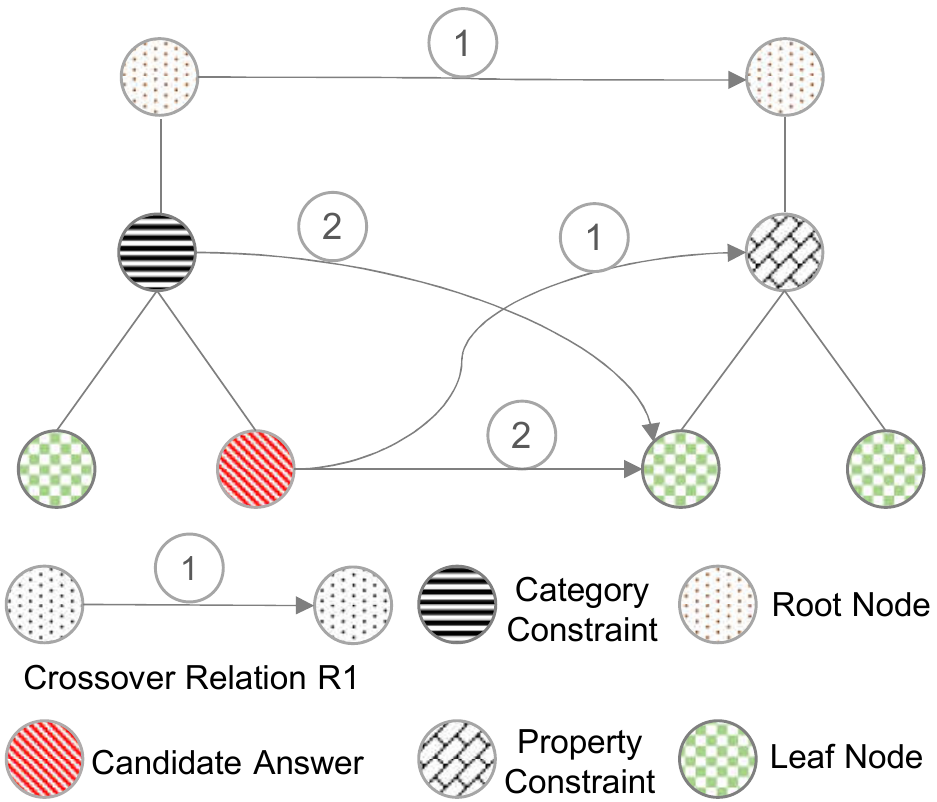}
    \vspace{-0.1cm}
    \caption {Diagram of Topological Structure of $SG$}
    \label{sgExample}
    \vspace{-0.3cm}
\end{figure}

\textbf{Topological Structure of Subgraph $SG$:} Fig.~\ref{sgExample} illustrates an example of $SG$.
The root node and leaf node refer to the most superclass node and most subclass node of $SG$, respectively.
The candidate answers are all the leaf nodes of $SG_{subj}$.

Assuming the red focus node in Fig.~\ref{sgExample} is a candidate answer for making decision~(to decide the candidate answer is a correct answer or not). The following two kinds of potential topological structures of the $CR$ can affect the decision-making.
$R1$ refers to the candidate answer or any superclasses of the candidate answer connected to $c_{obj}$ or any superclasses of $c_{obj}$ by the predicate.
$R2$ indicates that the candidate answer or any superclasses of candidate answer connect to any subclasses of $c_{obj}$ by the predicate.

\textbf{Predicate Similarity:}
There are many different ways to express the same meaning and opposite meaning.
Therefore, the predicates of $CR$ also affect the decision-making.
To measure the semantic relevance between two predicates, a technique named \emph{word2vec}~\cite{mikolov2013distributed,xu2016improve,liu2017revisit} is adopted.
It assumes that words appear in similar context may have similar meanings~\cite{harris1954distributional} and transforms the words or phrases into the form of continuous vectors~\cite{mikolov2013distributed}.
Then words or phrases can do relevance computation by distance measurement of vector~(here we use cosine similarity).

\textbf{Features Generations:}
In order to estimate the candidate answers, sufficient evidence need to be gathered to determine whether it surpasses the positive or negative criteria.
In general, the weights of positive and negative evidence are not equal.
A few negative evidence may lead to negative decisions, whereas many positive evidence are needed to make a positive decision.

%
%

\begin{table}[htb]
\centering
\caption{Features to Represent the Candidate Answer}
\label{features_DeveloperBot}
\begin{tabular}{@{}lll@{}}
\toprule
\# & Name            & Gloss                                                                                                                                                                                                                             \\ \midrule
1  & p\_r1\_pdictSim & \begin{tabular}[c]{@{}l@{}}If positive $R1$ does not exist, the value\\ takes 0. Otherwise, takes the maximum value\\ of predicates similarity.\end{tabular}
\\ \cdashline{1-3}[0.8pt/2pt]
2  & p\_r2\_pdictSim & \begin{tabular}[c]{@{}l@{}}If positive $R2$ does not exist, the value\\ takes 0. Otherwise, the value takes the sum\\ of predicates similarity of all $R2$ divide the\\ number of subclasses of property constraint.\end{tabular}
\\ \cdashline{1-3}[0.8pt/2pt]
3  & n\_r1           & \begin{tabular}[c]{@{}l@{}}If negative $R1$ does not exist,\\ the value takes 0. Otherwise, takes 1.\end{tabular}                                                                                                                 \\ \cdashline{1-3}[0.8pt/2pt]
4  & n\_r2           & \begin{tabular}[c]{@{}l@{}}If negative $R2$ does not exist, \\ the value takes 0. Otherwise, takes 1.\end{tabular}                                                                                                                \\ \bottomrule
\end{tabular}
\vspace{-0.6cm}
\end{table}

We synthesize the topological structure and predicate similarity into four features as shown in Tab.\ref{features_DeveloperBot}.
In these four features, feature \#1 represents the influence of $R1$.
If $R1$ does not exist, the value of feature \#1 takes 0, otherwise takes the maximum value of predicates similarity of all $R1$: $p\_r1\_pdictSim = max(pdictSim\_R1_i)$,
where $i = (1,2,..., n)$ and $pdictSim\_R1_{i}$ is the predicates similarity of $i$th $R1$, and $n$ is the number of $R1$.

Feature \#2 represents the influence of $R2$.
If $R2$ does not exist, the value takes 0.
Otherwise, the more $R2$ in $CR$, the higher the probability that the candidate answer meets the property constraint.
Therefore, the value of feature \#2 $p\_r2\_pdictSim = \frac{1}{mn}\sum_{i=1}^{m}\sum_{j=1}^{n}pdictSim\_R2_{ij}, i = (1,2,..., m), j = (1,2,..., n)$,
where $pdictSim\_R2_{ij}$ is the predicates similarity of $i$th leaf nodes of $c_{subj}$ connected to $j$th subclass of $c_{obj}$.
$m$ is the number of leaf nodes of $c_{subj}$ and $n$ is the number of subclasses of $c_{obj}$.
Take the Fig.~\ref{sgExample} as an example, here $m = 2$, and $n = 2$, assume the value of $pdictSim\_R2_{21} = 1$ and $pdictSim\_R2_{22} = 0.5$, the value of feature \#2 is $\frac{(1 + 0.5)}{4} = 0.375$.
For the feature \#3 and feature \#4, if negative $R1$ or $R2$ do not exist, the value takes 0, otherwise, takes 1.

\subsubsection{Final Answer and Explanation Generation}\label{answerGeneration}
\emph{BotReasoning} extracts the layer information of $QG$ and the answer reasoning will be performed from the outer layer to the inner layer.
The reasoning order of the constraint quads in same layer is determined by random.
Once the answers $AS = (as_1, as_2,...,as_n)$, where $n \geq 0, n \in \mathbb{Z}$ of a constraint quads $currentCQ$ are found, \emph{BotReasoning} will replace the corresponding object constraint in the inner layer by these answers and prune the $currentCQ$ from $QG$.
\emph{BotReasoning} will generate $n$ pruned $QG$, and continues the reasoning process for every $QG$.
If there are multiple constraints in the same layer and multiple answer sets are deduced, the union of these sets is computed.
These iterations will terminate until the final answers and explanations of the most inner layer of every $QG$ are achieved.

The corresponding explanations~(reasoning subgraph and answer confidence) are extracted following the cognitive process of \emph{DeveloperBot}.
\emph{BotPerception} records the entities identified in a query to the reasoning subgraph, these are also the initially activated nodes~($node_{subj}$ and $node_{obj}$) for spreading activation in the knowledge graph.
After every constraint reasoning, \emph{BotReasoning} takes all the $CR$ and the relational paths from $node_{subj}$ or $node_{obj}$ to any nodes of $CR$ into the reasoning subgraph of previous iteration~(if any).
Therefore, the final reasoning subgraph involves the information of whole cognitive process including \emph{BotPerception}, \emph{BotPlanning} and \emph{BotReasoning}.

The answer confidences $AC = (ac_1, ac_2,...,ac_m)$, where $m \geq 0, n \in \mathbb{Z}$ for $currentCQ$ are computed by the decision-making algorithm multiplicative the answer confidence from outer layer.
During the answer reasoning, the confidences $ac_i$ will be multiplicative to every confidence of the inner layer.
The confidence of an answer reasoned from multiple constraints in the same layer is obtained by multiplying the answer confidences of all the constraints in the same layer.


\subsection{\emph{BotResponse}: UI of \emph{DeveloperBot}}\label{BotResponse}
\begin{figure}[htb]
\vspace{-1.5mm}
\begin{center}
\includegraphics[width=0.46\textwidth,draft=false]{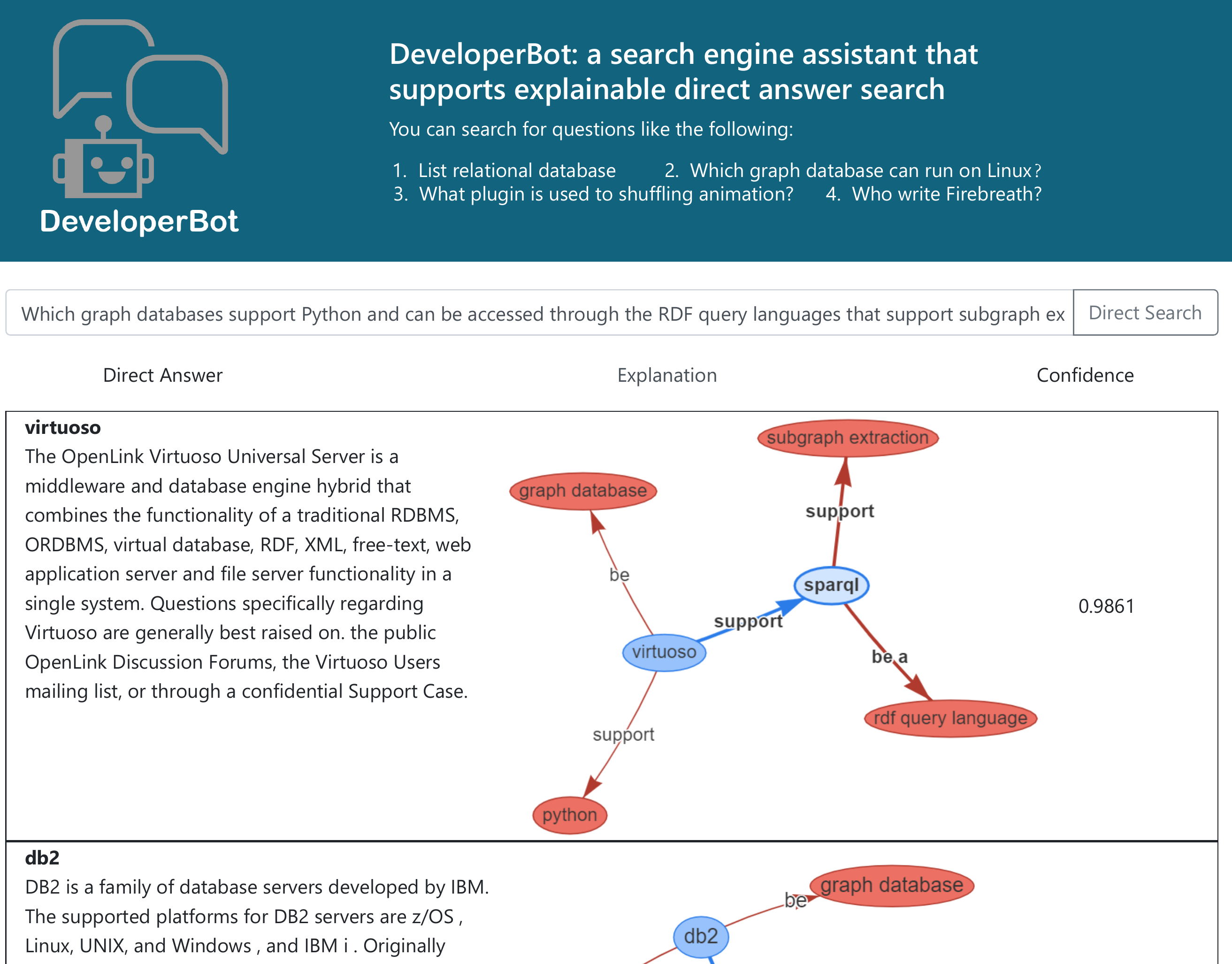}
\vspace{-0.1cm}
\caption{ \emph{BotResponse}: the User Interfaces of \emph{DeveloperBot}}
\label{ui}
\end{center}
\vspace{-0.55cm}
\end{figure}

A proof-of-concept prototype of \emph{DeveloperBot} is implemented as shown in Fig.~\ref{ui}.
The user can enter their query at the input bar and press the button ``Direct Search''.
The application can return a list of direct answers, brief introductions and their corresponding explanations.
In the reasoning subgraph, the entities identified by the \emph{BotPerception} module are represented as red colour.
The answers reasoning by \emph{BotReasoning} are presented as graded blue that shows how the \emph{BotPlanning} orders the constraints.

This reasoning subgraph explains ``why'' ``virtuoso'' is the answer: virtuoso is a graph database, virtuoso supports and sparql, and sparql is a rdf query language that supports subgraph extraction.
It also explain ``how'' ``virtuoso'' is achieved: firstly, \emph{DeveloperBot} identifies the four entities ``graph database'', ``python'', ``rdf query language'' and ``subgraph extraction''.
Secondly, it gets the answer ``sparql'' of constraint ``rdf query language support subgraph extraction''.
Finally, \emph{DeveloperBot} searches the ``graph database'' supported both python and sparql.
And the confidence of answer ``virtuoso'' is 0.9861.


\section{Experiment}
In this section, a series of experiments are conducted to answer the following three research questions of \emph{DeveloperBot}:
\begin{itemize}
  \item \textbf{RQ1}: How is the performance of different decision-making algorithms in estimating the correct answers?
  \item \textbf{RQ2}: Whether \emph{DeveloperBot} can improve the performance of developers during answer search? If the performance is different under different task complexity?
  \item \textbf{RQ3}: How is the performance of the different explanations and how these explanations affect user behavior during answer search?

\end{itemize}

\subsection{Data Acquisition}
\textbf{Knowledge Graph:}
The original knowledge graph is proposed by paper~\cite{zhao2017hdskg}.
This knowledge graph involves the general knowledge of the software engineering domain and contains 44800 concepts, 9660 unique verb phrase and 35279 relation triples.
Then, this knowledge graph is cleared up and augmented by the synonyms, expertise of domain experts, NLP techniques and partial knowledge of papers~\cite{zhaohdso,li2018improving}.
After that, the final knowledge graph acquires 39022 concepts, 8173 unique predicates and 35938 relation triples.
These knowledge cover the popular programming languages~(e.g., Java, Javascript, Python, etc.), frameworks~(e.g., Flex, Django, etc.), database~(e.g., neo4j, Mysql, etc.), tools~(e.g., d3, MSBuild, etc.) of software engineering domain.

\textbf{Training Corpus of Word Embedding:}
The tag wiki of Stack Overflow is used as the training corpus.
Tag wiki is not only well maintained, but also have standard grammatical structure and comprehensive technical coverage.
Furthermore, the semantic context of tag wiki matches to the knowledge graph we used well.
As a result, we acquire 20539 documents, then split 78466 sentences from the documents as training corpus of word embedding.

\subsection{Experimental Setup and Involved Tools}
In the subgraph search process of this experiment, the initial activation values of all node are set to zero.
Here the active threshold $AT$ is set to 0.8 and the decay factor $DF$ is set to 0.85.
For starting the activation, the activation values of linked nodes of subject constraint and object constraint will be initialized to greater than $AT$.
The maximum number of iterations is set to 30.
DNN model used in this experiment is a 3 layers classifier whose hidden layer is set to [10, 20, 10].
For the decision-making process of Bayesian Decision Theory, $p(\omega_i = 0) = 0.15, p(\omega_i = 1) = 0.85$.

The \emph{Stanford CoreNLP} is used for NLP markup, dependency parse~\cite{manning-EtAl:2014:P14-5} and named entity recognition.
The word2vec library of \emph{Gensim} is used to compute word embedding~\cite{rehurek_lrec}.
The \emph{nltk} library is used to do the VP, NP chunking~\cite{bird2006nltk}.

\subsection{Evaluation}
In this section, quantitative evaluation~(objective) and a user study~(subjective) are conducted to evaluate the performance of \emph{DeveloperBot}.

\subsubsection{Quantitative Evaluation}
In the quantitative evaluation, we invited five developers to ask some multi-constraint natural language questions they interested.
We require that these questions should be answered with direct answers.
These questions mainly cover factual questions such as Who, What, Which, and List.
Finally, the five developers proposed 160 eligible questions.
Then they asked to use computer to search their answers for every question.
Then they discuss together about the final answers for every question.
51 of the questions were dropped because they had no answer, or their answers could not reach a consensus.
Then the 109 ground truth questions and the 1819 candidate answers generated.

To evaluate the decision-making process, the balanced accuracy, precision, recall, f1-score and confidence mean square error~(MSE) are used as evaluation metrics~\cite{david2powers}.
The average result of the 10 times cross-validation will be taken as the final results of different metrics.
Here the MSE measures the average squared difference between the estimated confidence values and the actual confidence values: $MSE = \frac{1}{n}\sum_{i=1}^{n}(1-answerConfidence_i)^{2}$, where $n$ is the number of answers, the 1 is the actual confidence values for correct answers.
The balanced accuracy returns the average accuracy per class: $Balanced Accuracy = \frac{1}{2}\left( \frac{TP}{TP + FN} + \frac{TN}{TN + FP} \right)$,
where $TP$, $TN$, $FP$, $FN$ are true positives, true negatives, false positives, and false negatives, respectively.

%
%

\subsubsection{User Study}
\textbf{Tools for comparison:}
In this user study, we will compare the information search performance of using Google  + \emph{DeveloperBot} with use only Google.
There are some direct answer search engines like Wolfram Alpha~(https://www.wolframalpha.com/) and ASK~(https://www.ask.com) which are similar to \emph{DeveloperBot}.
But these tools can hardly answer specific questions of software engineer domain, so it makes no sense to compare with them.
Furthermore, this paper mainly focuses on a general method to answer questions and provide explanations by a knowledge graph.
It can be extended to many different areas, situations and questions coverage, which depend on the knowledge graph used.

\textbf{Participants:}
In the user study, 24 developers from different backgrounds were recruited.
They are made up of the employees, undergraduate students, PhD students, professors from different universities and professional developers from IT companies, etc.
Their programming proficiency distributions are 12.5\%, 75\% and 12.5\% for expert, competent and beginner, respectively.
The participants also have diverse programming background and their programming languages involved Python, Java, Javascript, SQL, C\#, etc.
In each proficiency level and background, the developers are halved to $G1$ and $G2$ randomly~(12 participants per group).
To exclude the effects of participants' programming level and fatigue, $G1$ and $G2$ will act as experimental group~(use Google  + \emph{DeveloperBot}) and control group~(use only Google) to finish each task circularly
Therefore, every developer has the opportunity to solve the simple or hard task by only Google or Google + \emph{DeveloperBot}.
They can also have a comprehensive comparison to the differences with or without the search engine assistant \emph{DeveloperBot}.

\begin{table}[htb]
\vspace{-2mm}
\caption{The Tasks for User Study}
\vspace{-1mm}
\centering
\begin{tabular}{@{}clc@{}}
\toprule
\multicolumn{1}{l}{Task No.} & Task Description                                                                                                                                                         & \multicolumn{1}{l}{Best Answer(s) \#} \\ \midrule
1                            & Who develops Hiawatha?                                                                                                                                                   & 1                                            \\ \cdashline{1-3}[0.8pt/2pt]
2                            & List data visualization library                                                                                                                                          & 5                                            \\ \cdashline{1-3}[0.8pt/2pt]
3                            & \begin{tabular}[c]{@{}l@{}}Which mobile operating system is\\ based on Linux kernel?\end{tabular}                                                                        & 4                                            \\ \cdashline{1-3}[0.8pt/2pt]
4                            & Which IDE support web2.0?                                                                                                                                                & 3                                            \\ \cdashline{1-3}[0.8pt/2pt]
5                            & \begin{tabular}[c]{@{}l@{}}Which graph databases support python\\ and can be accessed through the\\ RDF query languages that support\\ subgraph extraction?\end{tabular} & 3                                            \\ \bottomrule
\label{tasks}
\end{tabular}
\vspace{-6mm}
\end{table}
\textbf{Task:}
In this study, we have chosen 5 answer search tasks~(represent as $Task1-Task5$) from the ground truth questions as shown in Tab.~\ref{tasks}.
The answer to these tasks cover celebrity of software engineering domain, library, operating system, IDE, and database, etc.
According to the preliminary test, the difficulty of these five tasks increases in order.
The participants need to evaluate all the answers given by Google or \emph{DeveloperBot} comprehensively, and select $x$ best answers according to the required number of every task.

\textbf{Procedure:}
This user study begins with a brief training of \emph{DeveloperBot} by the experimenter.
After finishing each task, participants need to fill their confidence to the answers they provided~(on 5-point likert scale with 1 being the least confident and 5 being the most confident).
A screencapture software is required to run throughout the whole process that allows us to analyze the participants' behavior after the experiment.
At the end, all the participants will be asked to fill in the System Usability Scale (SUS) questionnaire~\cite{brooke1996sus} and do an open discussion focusing on their options about the different features of \emph{DeveloperBot}.

\subsection{Results and Results Analysis}

\subsubsection{Result: performance of different decision-making algorithms~(\textbf{RQ1})}

\begin{figure}[htb]
\vspace{-4mm}
\begin{center}
\includegraphics[width=0.48\textwidth,draft=false]{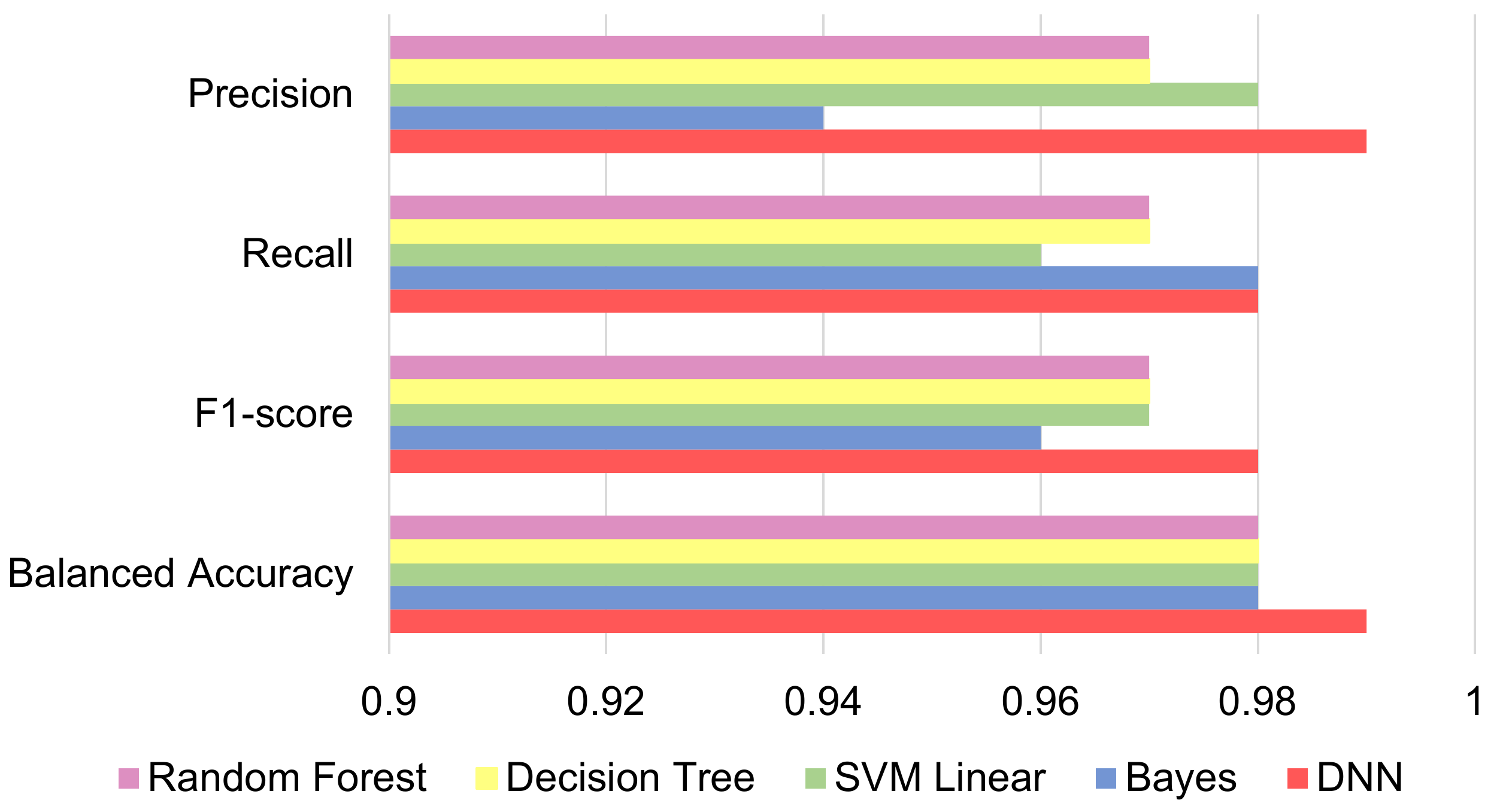}
\vspace{-1mm}
\caption{Performance of Different Decision-making Algorithms}
\label{performance_dm}
\end{center}
\vspace{-5mm}
\end{figure}

Here we will present the results of quantitative evaluation and answer the \textbf{RQ1}.
Fig.\ref{performance_dm} shows the performance of different decision-making algorithms including Random Forest, Decision Tree, SVM Linear, Bayes~(Bayesian Decision Theory), and DNN.
With the accurate representation of candidate answers by the four features extracted by our algorithm, any decision-making algorithm can get more than 0.98 balanced accuracy.
Compared with another algorithms, DNN achieves the highest precision 0.98, recall 0.98 and f1-score 0.99, respectively.
From Fig.\ref{performance_dm}, we can infer that some decision-making algorithms can achieve high precision, but at the cost of a low recall, like SVM Linear.
The Bayesian Decision theory obtains a high recall, but the precision is low.
In the future research, we might investigate the employment of other state-of-art algorithms, such as negative correlation learning~\cite{chen2010multiobjective,chen2009regularized}, statistical learning for probabilistic outputs~\cite{lyu2019multiclass,chen2013efficient}, Bayesian inference~\cite{chen2009probabilistic}, etc.

\begin{figure}[htb]
\begin{center}
\includegraphics[width=0.35\textwidth,draft=false]{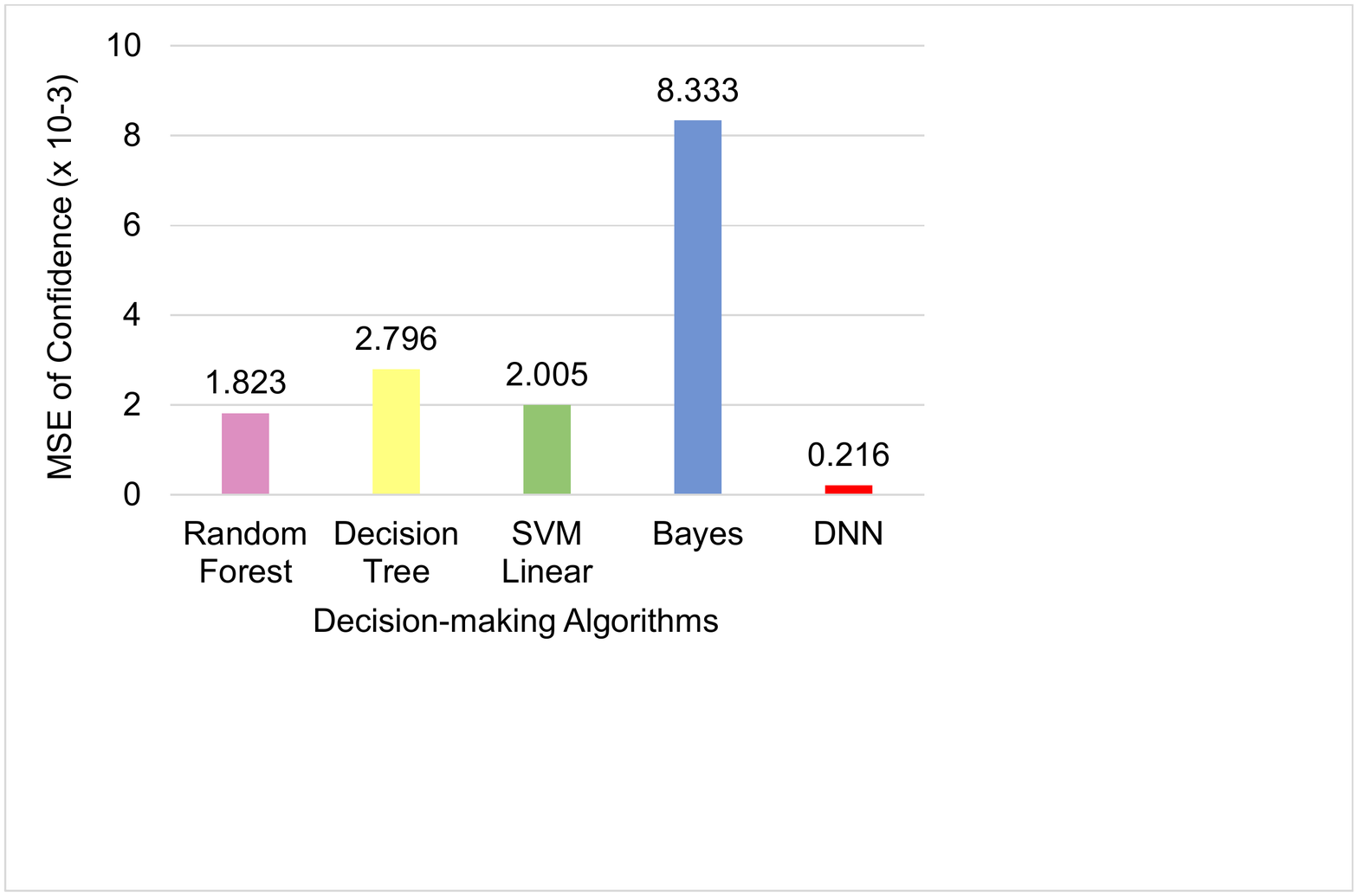}
\vspace{-0.1cm}
\caption{Answer Confidence MSE}
\label{confidence_dm}
\end{center}
\vspace{-0.9cm}
\end{figure}

Fig.\ref{confidence_dm} is the MSE of confidence for different decision-making algorithms.
It shows that DNN achieves the lowest MSE for the estimation of answer confidence.
Bayesian Decision Theory gets the highest MSE.
The estimation of answer confidence is within the acceptable range for all decision-making algorithms.

To explore the effects of different features, we train 3 DNN models using different features.
These 3 models use all features, only predicate similarity feature and only topological structure feature, respectively.
The only predicate similarity feature is computed by the maximal predicate similarity of $R1$, if it exists.
Only topological structure features involve 4 features to represent the existence~(take the value 1) or nonexistence~(take the value 0) of $R1$, $R2$, $R3$ and $R4$, respectively.

\begin{figure}[htb]
\vspace{-3mm}
\begin{center}
\includegraphics[width=0.42\textwidth,draft=false]{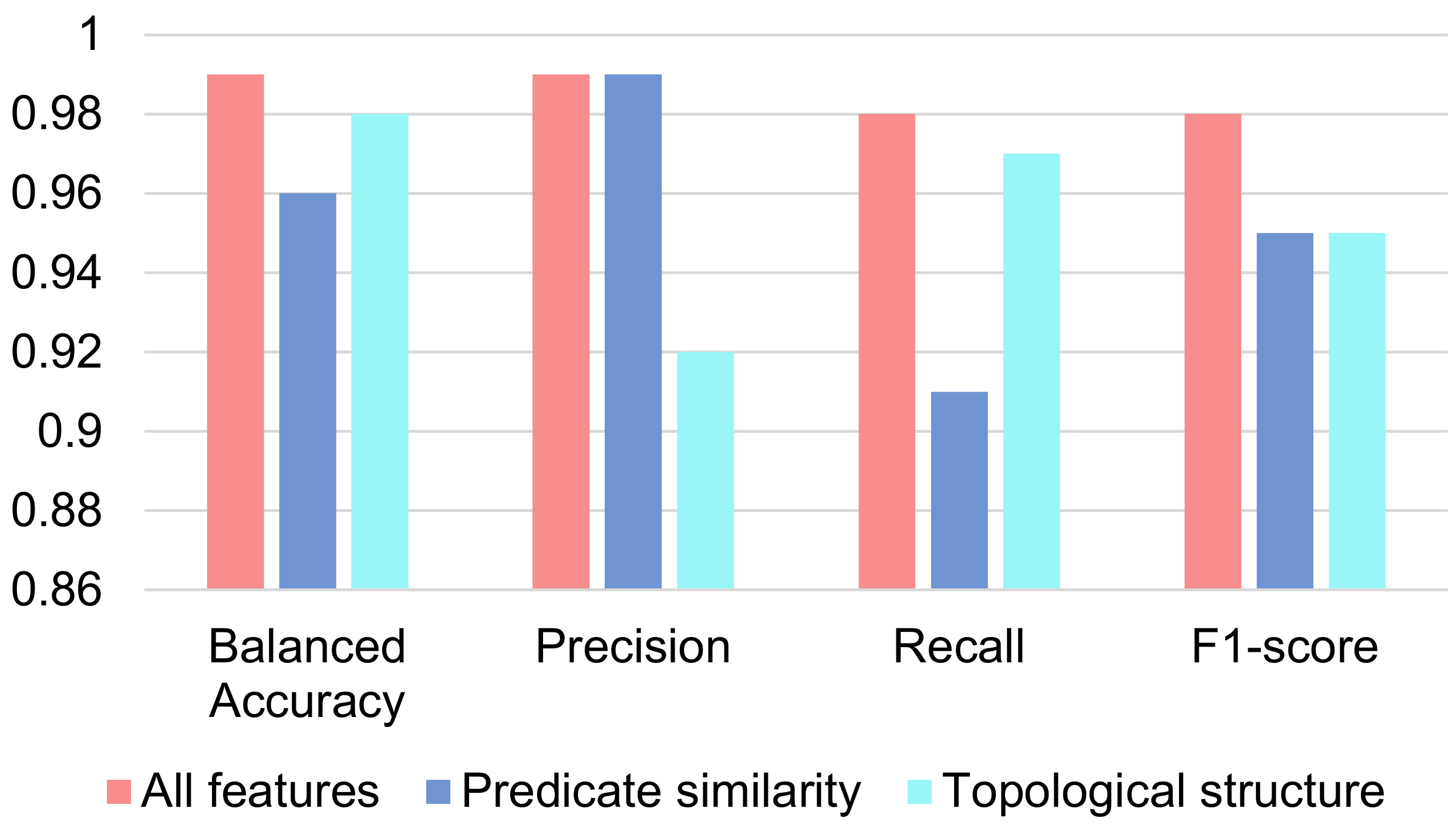}
\vspace{-1mm}
\caption{Performance of DNN with Different Features}
\label{features_performance}
\end{center}
\vspace{-5mm}
\end{figure}

As shown in Fig.\ref{features_performance}, the decision-making algorithm with all features performs the best for all the metrics.
This figure also indicates that the predicate similarity contributes to the precision significantly.
Since with only predicate similarity feature, the precision almost can achieve as same as all features.
But only predicate similarity feature obtains very low recall.
These results show that the predicate similarity is a good feature in terms of identifying the right answers, but without the topological structure, a lot of right answers are missed.
\begin{figure}[htb]
\vspace{-0.3cm}
\begin{center}
\includegraphics[width=0.35\textwidth,draft=false]{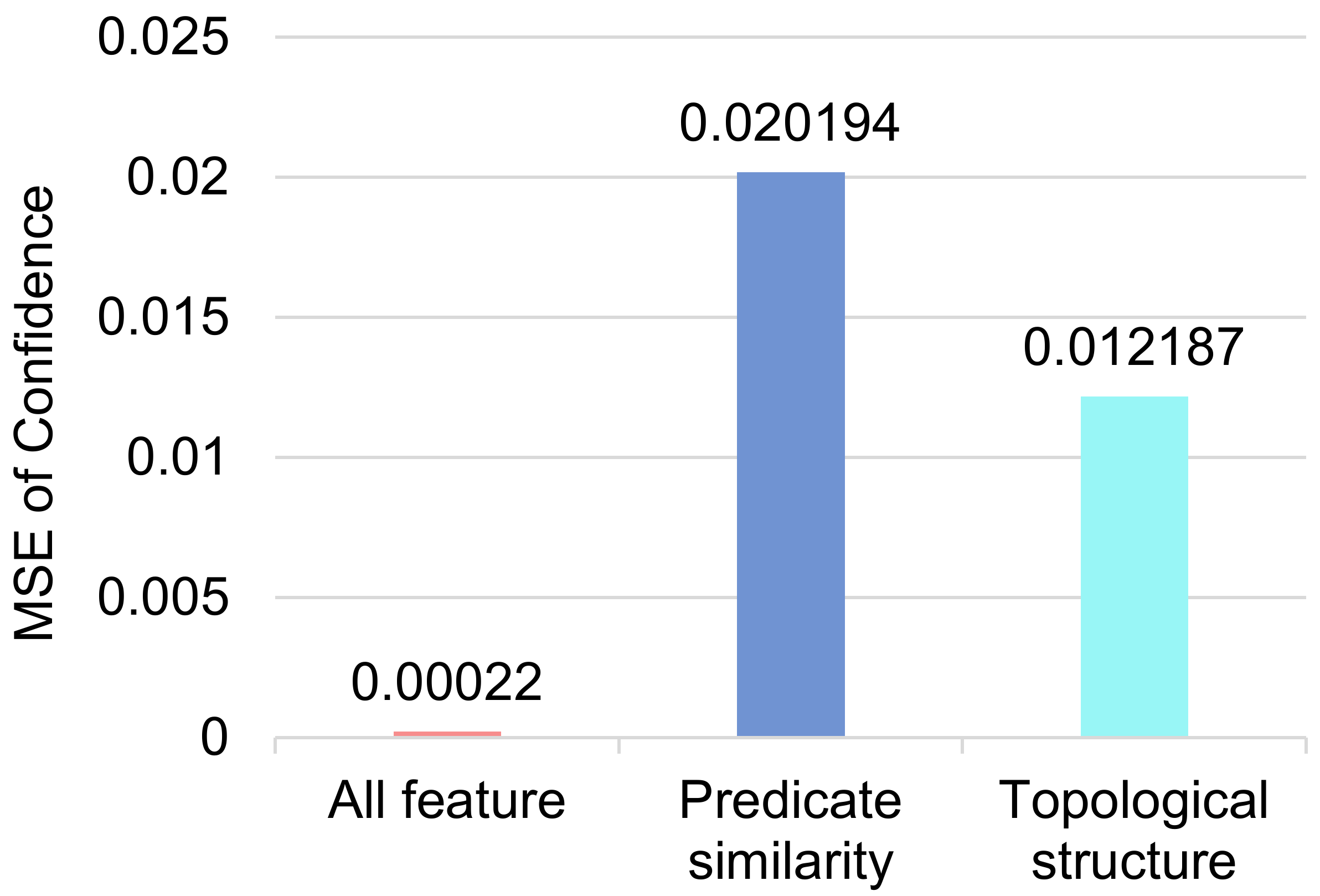}
\vspace{-0.1cm}
\caption{MSE of Confidence with Different Features}
\label{features_confidence}
\end{center}
\vspace{-0.5cm}
\end{figure}

Fig.\ref{features_confidence} indicates that the decision-making algorithm can obtain the lowest MSE by using all features.
Compared with the topological structure, predicate similarity achieves higher MSE.
From the above exploration, we can infer that both predicate similarity and topological structure we design for decision-making are useful for identifying the right answer from the candidate answers.

\subsubsection{Result: improve the performance of developers during answer search~(\textbf{RQ2})}
\begin{figure}[htb]
\begin{center}
\includegraphics[width=0.43\textwidth,draft=false]{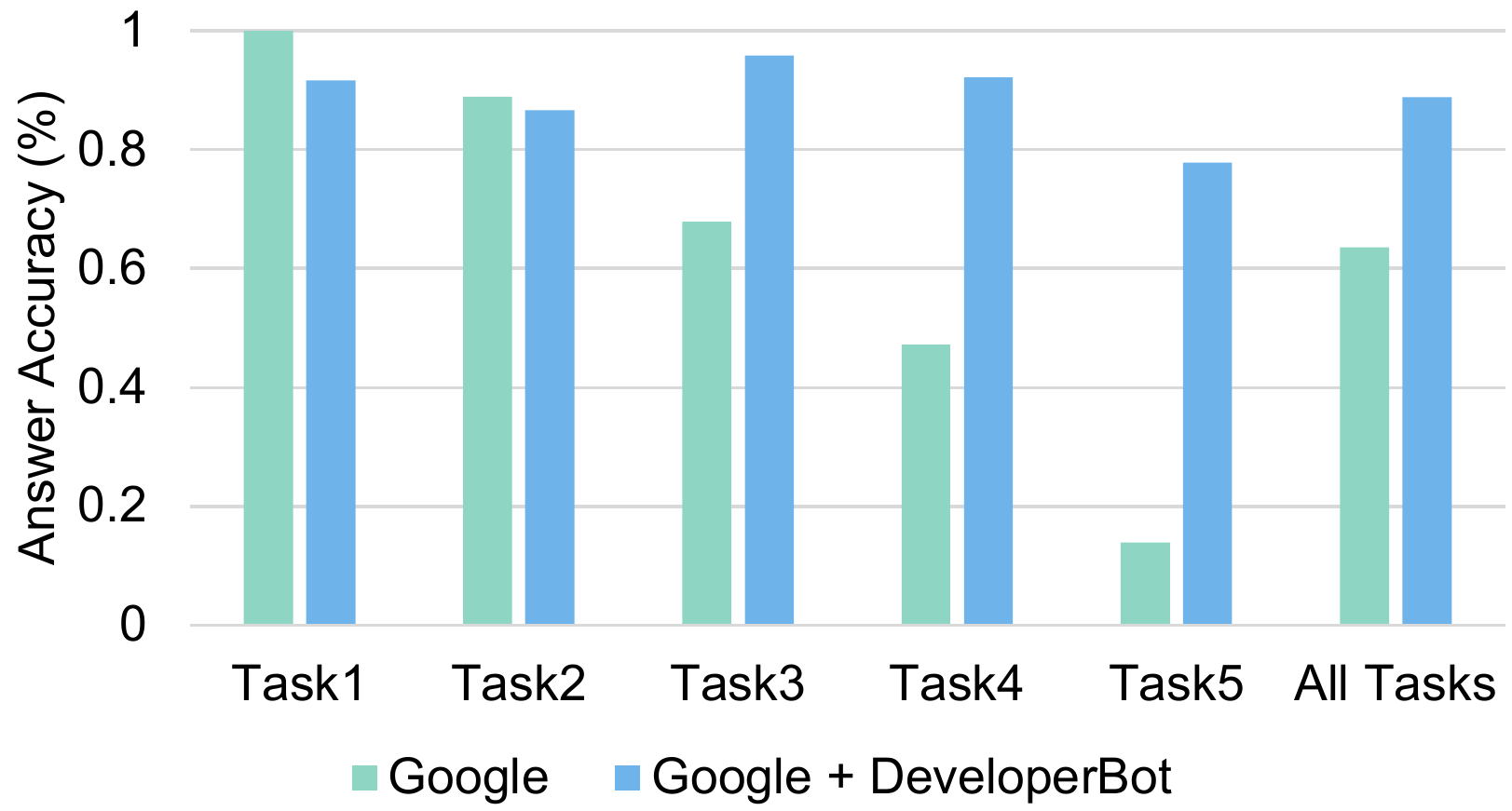}
\vspace{-0.1cm}
\caption{Answer Accuracy for Each Task}
\label{accuracy_userStudy}
\end{center}
\vspace{-0.9cm}
\end{figure}
Fig.\ref{accuracy_userStudy} is the result of the average answer accuracy of the two groups.
It shows that for the task 1-2, both groups can finish the tasks with high accuracy~(above 0.86).
The answer accuracy of the participants who use only Google even reaches 1 on task 1.
These prove our pre-experiment conjecture: Google has a very good performance in simple information search tasks.
For the task 3-5, the experimental group can finish with higher answer accuracy.
With the complexity of tasks increasing, the answer accuracy of the control group gradually decreases compared with the experimental group which maintains at a high level~(above 0.78).
For task 5, the difference in answer accuracy even reaches 0.64.
These fully illustrate that \emph{DeveloperBot}, as a search engine assistant, can significantly improve the answer accuracy during search close-end questions, especially in complex search tasks.

\begin{figure}[htb]
\vspace{-0.4cm}
\begin{center}
\includegraphics[width=0.45\textwidth,draft=false]{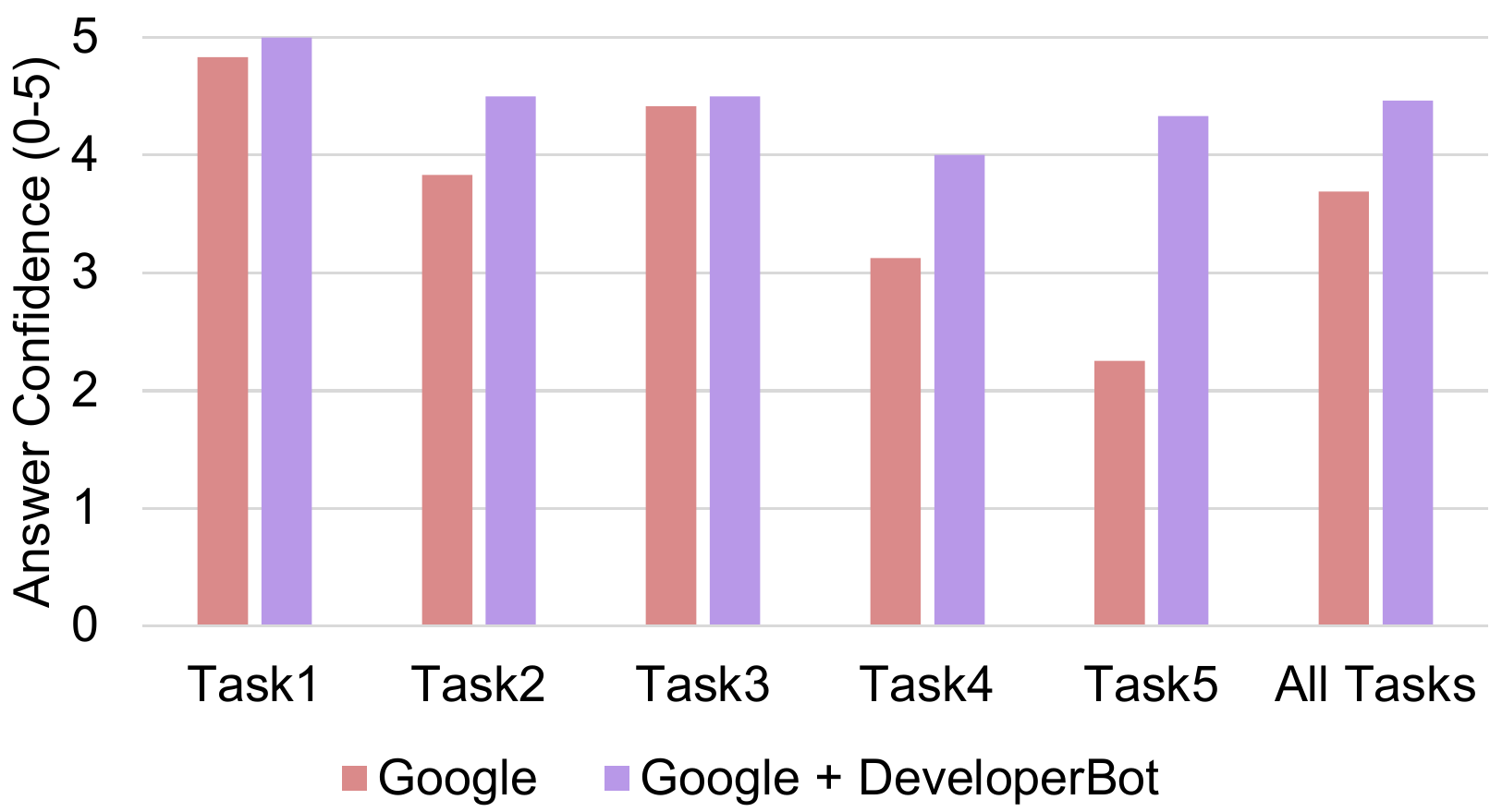}
\vspace{-0.2cm}
\caption{Participants' Confidences to Answers for Each Task}
\label{confidence_userStudy}
\end{center}
\vspace{-0.57cm}
\end{figure}

\begin{figure}[htb]
\vspace{-0.35cm}
\begin{center}
\includegraphics[width=0.45\textwidth,draft=false]{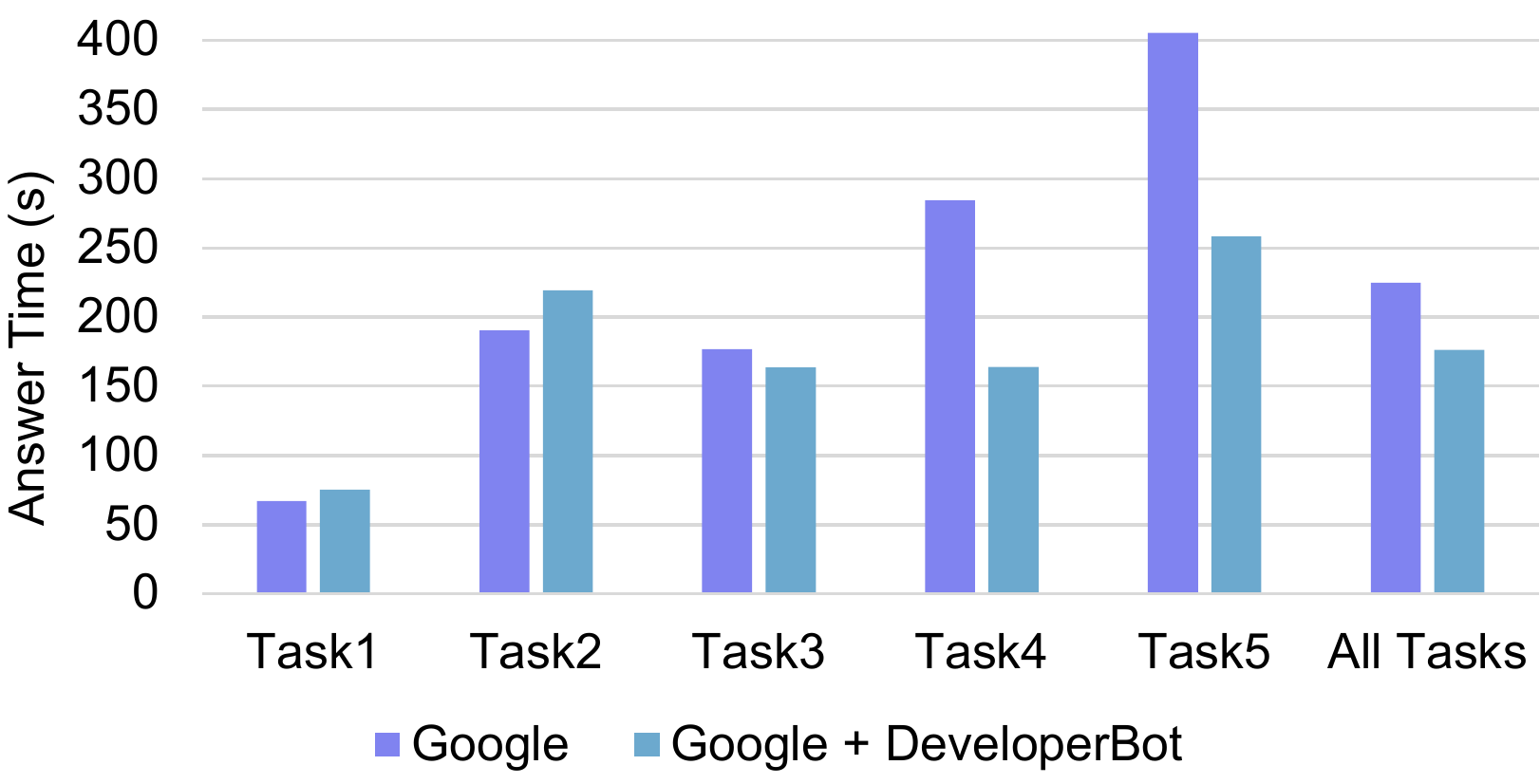}
\vspace{-0.15cm}
\caption{Answers Search Time for Each Task}
\label{time_userStudy}
\end{center}
\vspace{-0.55cm}
\end{figure}

Fig.\ref{confidence_userStudy} is the result of participants' average confidences to their answers~(called participants' confidences).
This figure indicates that for tasks 1-3, the participants' confidences of control group are almost same as experimental group.
For tasks 4-5, the participants' confidences~(4 and 4.3 for task 4 and task 5, respectively) of the experimental group show an overwhelming advantage.
With the complexity of tasks increasing, the difference of participants' confidences of two groups shows a clear increasing trend except task 3.
%
According to the observation of the screen recording and results of the open discussion, this is because there is a web page that lists some direct answers of task 3, and each answer has corresponding text explanation.
The participants' confidences of control group increases by reading these text explanation.
But this is at the cost of increasing the time used on task 3 as shown in Fig.~\ref{time_userStudy}.
These indicated that for the easy tasks, the control group was more confident in their answers because they were straightforward and did not require reasoning or summarization.
For the complex tasks, despite spending a long time to search answers, participants still feel less confident to their answers because they may lose during they try to summarize and reason the answers through many web pages.
Therefore, as a search engine assistant, \emph{DeveloperBot} can improve the participants' confidences by providing explanations of answers.


Fig.\ref{time_userStudy} is the result of answers search time.
It shows that, in addition to task 3, the answers search time of the control group increases significantly with the complexity of the tasks increasing.
With the assistant of \emph{DeveloperBot}, the answer search time of the experimental group for task 5 is 258s compared with 405s of the control group.
These show that using \emph{DeveloperBot} as search engine assistant can significantly reduce the participants' answer search time especially for the complex answer search tasks that take a long time to solve with only Google.
The open discussion also reflects consistent results for the above conclusion.
The special performance of Google group at task 3 also illustrates that direct answers and the explanation~(even text) can significantly reduce the search time and improve the participants' confidences to their answers at the same time.

\subsubsection{Result: comments to explanations and behavior changes~(\textbf{RQ3})}

We will evaluate the explanation and answer the RQ3 from three ways: the quantitative scoring~(0-5) of participants in the questionnaires, the information obtained from open discussions, and our observation of screen recording.

\begin{figure}[htb]
\begin{center}
\includegraphics[width=0.5\textwidth,draft=false]{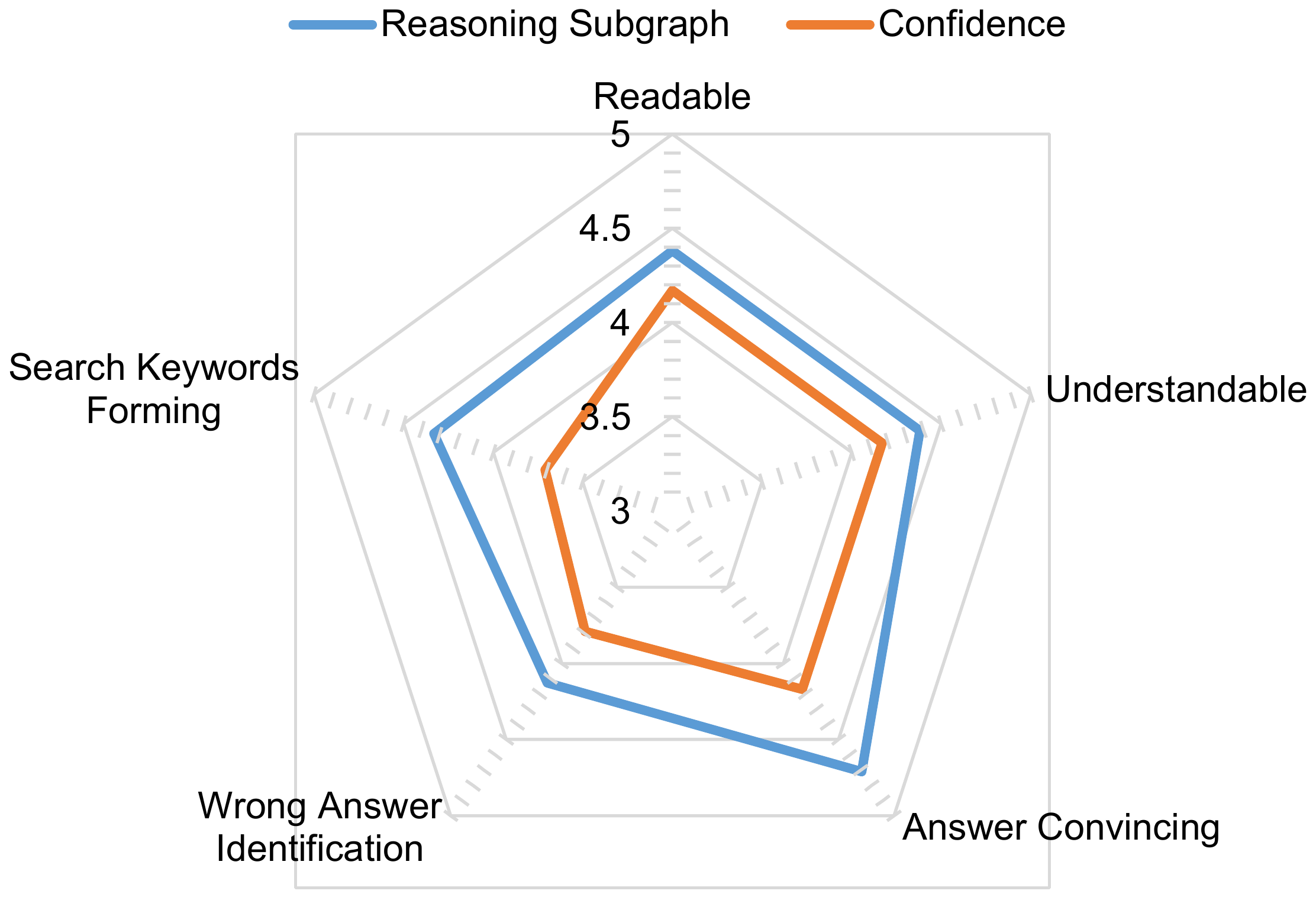}
\vspace{-0.5cm}
\caption{Quantitative Scoring of the Explanations}
\label{radar_resSubgConf}
\end{center}
\vspace{-0.8cm}
\end{figure}

The results of the quantitative scoring of explanations are shown in the radar graph Fig.~\ref{radar_resSubgConf}.
Five indexes are adopted in this experiment to evaluate the quality of the explanation.
They are Readable, Understandable, Answer Convincing, Wrong Answer Identification and Search Keywords Forming~\cite{confalonieri2019makes,hoffman2018metrics}, where Readable and Understandable mean the explanation is legible, easy to read and understand.
Answer Convincing means this explanation can make the direct answer more convincing.
Wrong Answer Identification is the index that evaluates if the explanation can help identify the wrong answer.
Search Keywords Forming shows the capacity of the explanation to help form better search keyword.

As shown in Fig.~\ref{radar_resSubgConf}, the overall scores of reasoning subgraph reach 4.13-4.71.
Compared with the other four indexes, the Answer Convincing achieves the highest score 4.71.
The ranking of indexes of Readable and Understandable are second and third, with average values of 4.38.
Even the Wrong Answer Identification achieves the lowest score, 4.13 is still acceptable.
The results demonstrate that the reasoning subgraph is a reasonable and high-quality explanation of direct answer and meet the desired design goals.

In addition, Fig.~\ref{radar_resSubgConf} also shows the overall scores of using confidence as an explanation.
Compared with the reasoning subgraph, the overall scores of confidence are relatively lower~(3.71-4.17).
The indexes Readable, Understandable and Answer Convincing achieve highest score 4.17.
The scores of Wrong Answer Identification and Search Keywords Forming is 3.79 and 3.71, respectively.
Even the overall scores of confidence are lower than reasoning subgraph, especially the Search Keywords
Forming and Wrong Answer Identification.
However, the good performance of indexes like readable, understandable and answer convincing indicate that confidences are good explanations to direct answers.


The results of the questionnaire and the contents of the open discussion show that, using the reasoning subgraph extracted following the cognitive process and answer confidence as explanations of the direct answers can significantly improve the developers' trust and adoption to the answers.
These explanations also assist the developers to understand the answers more deeply, improve the answer accuracy and form better search keywords.

\section{Conclusion}
The developers have a need for complex information search that provide specific direct answers.
The current search engines are weak at the reasoning capacity for complex closed-end questions.
In order to address these issues, in this paper, a brain-inspired search engine assistant named \emph{DeveloperBot} is proposed.
\emph{DeveloperBot} aligns to the cognitive process of human and has the capacity to answer complex queries with good explainability by learned knowledge.
The experimental results show that the novel features of the subgraph can estimate the answers and answer confidences with high accuracy.
The results of user study show that compared with using only Google, with the assistance of \emph{DeveloperBot}, users can find answers faster and with more accuracy.
In addition, with the explanations extracted following the cognitive process, \emph{DeveloperBot} can significantly improve the developers' trust and adoption to the answers.
These explanations also assist the developers to understand the answers more deeply, improve the answer accuracy and form better search keywords.
Furthermore, for relatively complex queries, with the assistance of \emph{DeveloperBot}, the search performance improvement of the developers is more significant.
In the future, we will augment the knowledge graph through developers' development behaviors to further improve the efficiency of information search
~\cite{zhao2018smart,li2018improving}.

\bibliographystyle{IEEEtran}
\bibliography{IEEEabrv,saner2017}

\end{document}